\documentclass[journal]{IEEEtran}
\usepackage[utf8]{inputenc}
\usepackage[T1]{fontenc}

\usepackage{amsmath}
\usepackage{amsfonts}
\usepackage{amssymb}

\usepackage{amsthm}
\usepackage[ruled,vlined]{algorithm2e}
\usepackage{mathtools}
\usepackage{tabularx}
\usepackage{graphicx}
\usepackage{subfigure}
\usepackage{enumerate}
\usepackage{float}
\usepackage{url}
\usepackage{verbatim}
\usepackage{stfloats}
\usepackage{bbding}
\usepackage{multirow}

\usepackage[normalem]{ulem}
\usepackage{times}
\usepackage{xcolor}
\usepackage{cancel}

\newcommand{\delete}[1]{{\bgroup\markoverwith{\textcolor{red}{\rule[0.5ex]{2pt}{0.4pt}}}\ULon{#1}}}
\newcommand{\deletefig}[1]{{\bgroup\markoverwith{\textcolor{red}{\rule[2.5ex]{2pt}{2.0pt}}}\ULon{#1}}}

\makeatletter
\let\NAT@parse\undefined
\makeatother

\definecolor{rblue}{rgb}{0,0.5,1}
\usepackage{hyperref}
\hypersetup{colorlinks=true,linkcolor=red,citecolor=rblue}

\usepackage{booktabs}
\usepackage{threeparttable}  
\usepackage{multirow}
\usepackage{pifont}  % for checkmark

\usepackage{xcolor}
\definecolor{revised_color}{HTML}{00008B}

\newcommand{\purple}[1]{\textcolor[RGB]{255,0,255}{#1}}
\newcommand{\green}[1]{\textcolor[RGB]{0,255,0}{#1}}

\definecolor{rblue}{rgb}{0,0.5,1}

\usepackage{cite}

\graphicspath{{pics/}}

\begin{document}
\title{LF-PGVIO: A Visual-Inertial-Odometry Framework for Large Field-of-View Cameras using Points and Geodesic Segments}
\author{Ze Wang, Kailun Yang\IEEEauthorrefmark{1}, Hao Shi, Yufan Zhang, Zhijie Xu, Fei Gao, and Kaiwei Wang\IEEEauthorrefmark{1}%
\thanks{This was supported in part by the National Natural Science Foundation of China (Grant No. 12174341), in part by Henan Province Key R$\&$D Special Project (2311111112700), in part by Hangzhou SurImage Technology Company Ltd., and in part by Hangzhou HuanJun Technology Company Ltd.}
\thanks{Z. Wang, H. Shi, Y. Zhang, and K. Wang are with the State Key Laboratory of Extreme Photonics and Instrumentation and with the National Engineering Research Center of Optical Instrumentation, Zhejiang University, Hangzhou 310027, China.}
\thanks{K. Yang is with the School of Robotics and the National Engineering Research Center of Robot Visual Perception and Control Technology, Hunan University, Changsha 410082, China.}
\thanks{Z. Xu is with the School of Computing and Engineering, University of Huddersfield, Huddersfield HD1 3DH, UK.}
\thanks{F. Gao is with the State Key Laboratory of Industrial Control Technology, Zhejiang University, Hangzhou 310027, China.}
\thanks{\IEEEauthorrefmark{1}Corresponding authors: Kaiwei Wang and Kailun Yang. (E-mail: wangkaiwei@zju.edu.cn, kailun.yang@hnu.edu.cn.)}%
}

\markboth{IEEE Transactions on Intelligent Vehicles, March~2024}%
{Wang \MakeLowercase{\textit{et al.}}: LF-PGVIO}

\maketitle

\begin{abstract}
In this paper, we propose LF-PGVIO, a Visual-Inertial-Odometry (VIO) framework for large Field-of-View (FoV) cameras with a negative plane using points and geodesic segments. The purpose of our research is to unleash the potential of point-line odometry with large-FoV omnidirectional cameras, even for cameras with negative-plane FoV. To achieve this, we propose an Omnidirectional Curve Segment Detection (OCSD) method combined with a camera model which is applicable to images with large distortions, such as panoramic annular images, fisheye images, and various panoramic images. The geodesic segment is sliced into multiple straight-line segments based on the radian and descriptors are extracted and recombined. Descriptor matching establishes the constraint relationship between 3D line segments in multiple frames. In our VIO system, line feature residual is also extended to support large-FoV cameras. Extensive evaluations on public datasets demonstrate the superior accuracy and robustness of LF-PGVIO compared to state-of-the-art methods. The source code will be made publicly available at \url{https://github.com/flysoaryun/LF-PGVIO}.

\end{abstract}

\begin{IEEEkeywords}
Visual-inertial-odometry, large-FoV cameras, curve segment detection, SLAM.
\end{IEEEkeywords}

\IEEEpeerreviewmaketitle

\section{Introduction}

\begin{figure}[!t]
%\vskip-1ex
\centering
\includegraphics[width=1.0\linewidth]{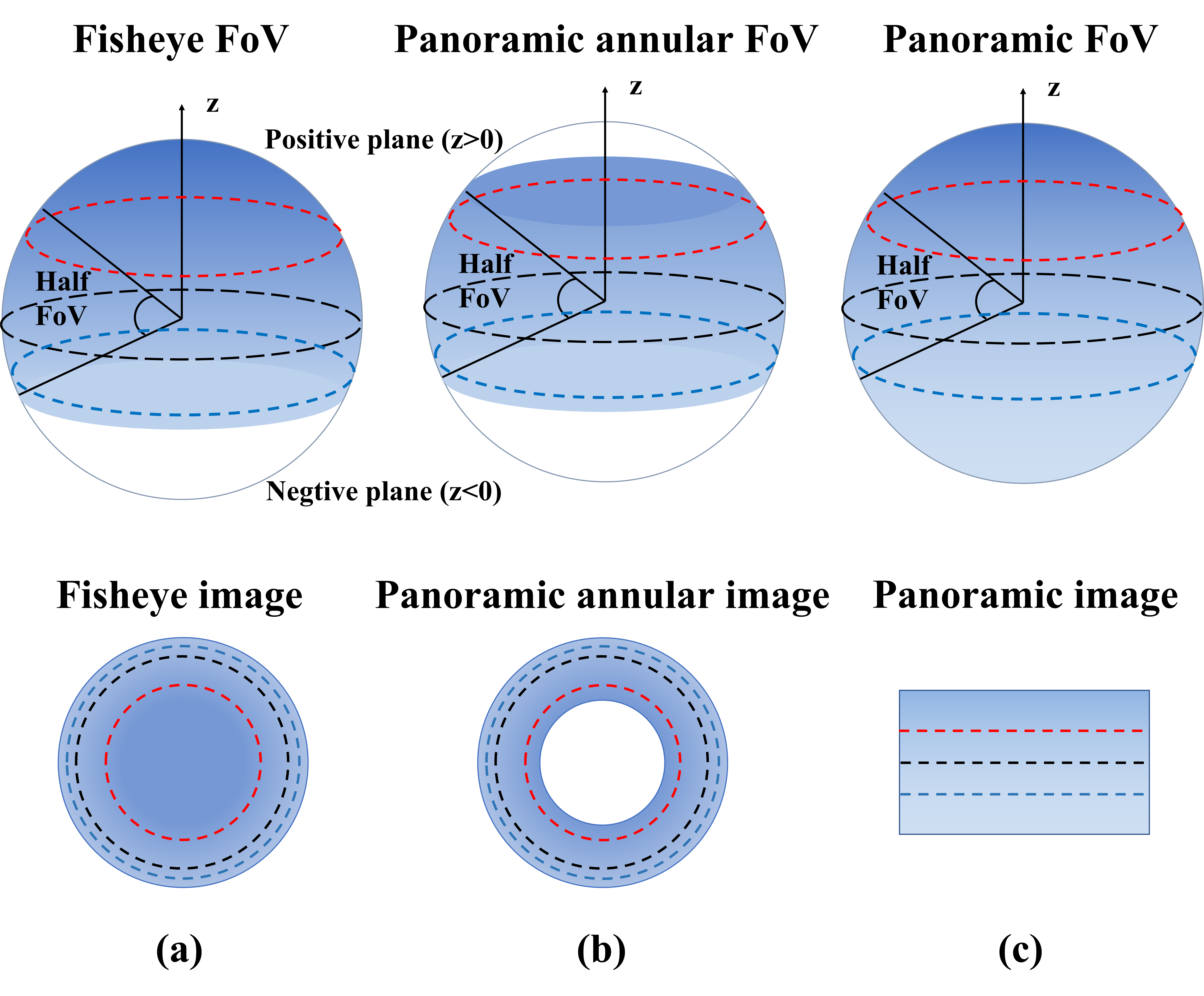}
%\vskip-1ex
\caption{Three typical omnidirectional images that may have a negative plane and the corresponding FoV. \textbf{(a)} Fisheye FoV and Fisheye image. \textbf{(b)} Panoramic annular FoV and Panoramic annular image. \textbf{(c)} Panoramic FoV and Panoramic image.}
\label{fig:negative_plane}
%\vskip-3ex
\end{figure}

\IEEEPARstart{T}{he} technology of Visual-Inertial-Odometry (VIO) plays a crucial role in accurately estimating the motion of intelligent vehicles and robots~\cite{nourani2009practical,zhou2022swarm,weinstein2018visual,chalvatzaras2022survey,hu2023nalo,huang2023mc_veo,pang2023structural,zhuoins2023_4drvo,shu2022multimodal_feature_constraint}.
Point-line VIO is particularly useful in scenarios where the environment has a lot of straight edges, such as indoor environments with many walls and corners or outdoor road environments with many buildings, zebra crossing, or road edges~\cite{gomez2019pl,pumarola2017pl,fu2020pl,lim2022uv,zhao2023visual}.
{Line extraction is a critical step in point-line odometry, which is a technique used to estimate the motion of a camera or vehicle by tracking the 3D positions of feature points and lines over time~\cite{li2021ulsd,suarez2022elsed}.
Traditionally, line extraction in point-line odometry~\cite{gomez2019pl,pumarola2017pl,fu2020pl,lim2022uv} involves first undistorting the image and then extracting lines on the pinhole image.
However, when the Field of View (FoV) of the camera is particularly large and there is a negative imaging half-plane FoV ($z<0$, see Fig.~\ref{fig:negative_plane}), it becomes unfeasible to undistort the image into a single pinhole representation encompassing the entire FoV. These limitations restrict the applicability of traditional line extraction methods in point-line odometry.}

To address the aforementioned issues and to unleash the potential of point-line odometry with large-FoV omnidirectional cameras, we have proposed a method to extract geodesic segments directly on the distorted image based on the camera model, without the need for undistortion.
Our method extracts omnidirectional curve segments from distorted images that are geodesic segments on the unit sphere, serving as necessary conditions to represent projections of 3D line segments. This approach allows for the extraction of line segments even in the presence of negative half-plane FoV, maintaining the continuity of extracted geodesic segments between frames. Furthermore, we partition the detected geodesic segments into multiple line segments according to the corresponding arc of the omnidirectional curve segment, obtaining straight-line sub-descriptors and combining them into a corresponding descriptor for matching. Our OCSD method alleviates the computational burden and has higher accuracy compared with other methods, such as RHT~\cite{torii2007randomized} and ULSD~\cite{li2021ulsd}. The OCSD method improves the robustness of point-line odometry, especially for large-FoV images. Our method enables more efficient and robust point-line odometry in practical applications.
Moreover, we put forward to seize the geodesic segments to construct line feature residual to solve the problem that traditional line segments cost residual~\cite{gomez2019pl,pumarola2017pl,fu2020pl,lim2022uv} only support positive half-planes, and define the line feature residual on the unit sphere instead of on the $z{=}1$ plane.

\begin{figure*}[t]
	\centering
	\includegraphics[width=1.0\linewidth]{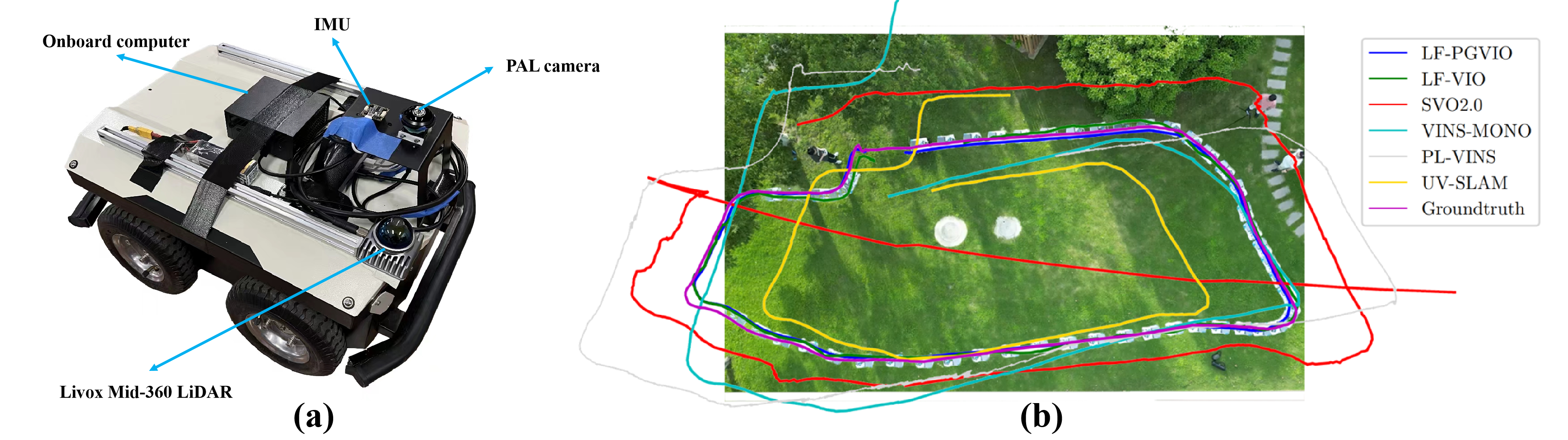}
	%\vskip-1ex
        \caption{$\textbf{(a)}$ Our car experiment platform with a Panoramic Annular Lens (PAL) camera, a Livox-Mid-360 LiDAR, an IMU sensor, and an onboard computer. $\textbf{(b)}$ Top view of trajectories of different algorithms and ground truth for the OD01 sequence in outdoor experiments. The car platform stacks images in a residual manner with a $0.5s$ interval on the first frame, and the trajectory aligns with the ground truth.} 
	\label{fig:car}
%\vskip-3ex
\end{figure*}

To address the scarcity of point-line visual-inertial odometry for large-FoV cameras, we propose the LF-PGVIO framework. 
In the front end of our framework, we introduce a novel approach for detecting omnidirectional curve segments and converting them into geodesic segments. In the back end of our framework, we define a new line feature residual on the unit sphere that allows for the recovery of 3D lines within the entire FoV. 
Meanwhile, our system can utilize 3D straight lines from the complete FoV to effectively improve our pose estimation.
To the best of the authors' knowledge, LF-PGVIO is the first visual-inertial odometry framework designed for large-FoV cameras using points and geodesic segments.

Extensive experiments are conducted on public datasets and real-world recordings.
On the TUM visual-inertial dataset~\cite{schubert2018tum}, we verify that the line feature residual with geodesic segments can effectively improve the VIO accuracy. On the public PALVIO dataset~\cite{wang2022lf,wang2023lf}, we conduct extensive experiments to verify the effectiveness of the algorithm compared with other state-of-the-art visual odometry and omnidirectional SLAM algorithms. 
The results show that our proposed Omnidirectional Curve Segment Detection (OCSD) method has a higher repetition rate and takes less time compared with other methods and the proposed LF-PGVIO has achieved higher accuracy compared to other algorithms on the challenging PALVIO dataset. 
We have further expanded the PALVIO dataset and recorded outdoor sequences to test our algorithm. 
The effect of the OD01 sequence captured via a mobile robot equipped with Panoramic Annular Lens (PAL) and IMU sensors is shown in Fig.~\ref{fig:car}, which proves the robustness and accuracy of our algorithm.

At a glance, we deliver the following contributions:

\begin{itemize}
\item We propose a novel OCSD method that extracts curve segments directly on distorted omnidirectional images with severe deformations, improving the efficiency and robustness of visual-inertial odometry.
\item To the best of the authors' knowledge, LF-PGVIO is the first visual-inertial odometry framework designed for large Field-of-View (FoV) cameras using points and geodesic segments. Our framework improves localization accuracy and stability in real-world scenarios. Our proposed framework effectively tackles the problem that large-FoV cameras cannot fully utilize line features projected onto the negative plane.
\item A comprehensive set of experiments on public datasets and real-world recordings demonstrates the effectiveness of our proposed OCSD method and the LF-PGVIO framework. Our LF-PGVIO can run in real time on mobile platforms.
\end{itemize}

\section{Related Work}
In this section, we provide an overview of advancements in line segment detection, large-FoV Simultaneous Localization and Mapping (SLAM), and point-line SLAM.
These discussed works demonstrate the potential of incorporating line features in visual-inertial odometry systems and offer various solutions for addressing the challenges posed by large-FoV cameras.

\subsection{Line Segments Detection}
The detection of line or curve segments in images has been developed for many years.
It is divided into two main categories: one based on traditional methods and the other based on deep learning methods.

Traditional methods, such as the Line Segment Detector (LSD)~\cite{von2012lsd} proposed by Von Gioi~\textit{et al.}, operate by detecting changes in intensity along a contour and using the information to identify line segments.
Almazan~\textit{et al.}~\cite{almazan2017mcmlsd} introduced a dynamic programming approach to line segment detection, which combines the advantages of perceptual grouping in the image domain and global accumulation in the Hough domain.
Su{\'a}rez~\textit{et al.}~\cite{suarez2022elsed} proposed Enhanced Line SEgment Drawing (ELSED). ELSED leverages a local segment-growing algorithm that connects gradient-aligned pixels in the presence of small discontinuities.

Recently, learning-based approaches for line segment detection have gained increasing attention due to their high accuracy and robustness in various scenarios.
Xue~\textit{et al.}~\cite{xue2020holistically} designed a fast and parsimonious parsing method for vectorized wireframe detection in an input image.
DeepLSD~\cite{pautrat2022deeplsd}, proposed by Pautrat~\textit{et al.}, combines traditional and learning-based approaches for line segment detection.
The method processes images with a deep network to generate a line attraction field before converting it to a surrogate image gradient magnitude and angle, which is fed into any existing handcrafted line detector.
{Torii~\textit{et al.}~\cite{torii2007randomized} proposed a Randomized-Hough-Transform (RHT) based method for great-circle detection on the sphere.
Pautrat~\textit{et al.} proposed SOLD2~\cite{pautrat2021sold2}, a learning-based method that jointly detects and describes line segments in an image in a single network without the need to annotate line labels.}
Li~\textit{et al.}~\cite{li2021ulsd} proposed a Unified Line Segment Detection (ULSD), which aims to detect line segments in both distorted and undistorted images.

However, these line detection methods all address straight line segments on pinhole images, and for images with large distortions, these straight lines will become multiple line segments, or even directly undetectable.
In this paper, we propose a camera model-based curve segment detection method that can detect ``straight line segments'' on large-FoV distorted omnidirectional images with severe deformations.

\subsection{Large-FoV SLAM}
Large-FoV SLAM has received a lot of attention due to its ability to obtain larger FoV and to perceive more of the surrounding environment, making localization more robust.
The large-FoV image acquisition is usually divided into two types: one is directly acquired by panoramic annular cameras~\cite{gao2022review,wang2022high_performance} or fisheye cameras~\cite{qian2022survey_fisheye,jakab2024surround_fisheye}, and the other is to obtain multiple images, and then combine them into one image according to the camera model. 
Panoramic annular cameras~\cite{jiang2022annular,jiang2023minimalist,yang2019pass} typically have a higher resolution within the large-FoV region, in particular on the negative half-plane, compared to fisheye cameras, but they do have blind spots.
For the first type of images, Wang~\textit{et al.}~\cite{wang2022pal_slam} proposed PAL-SLAM, a feature-based system for a panoramic annular camera that overcomes the limitations of traditional visual SLAM in rapid motion.
Chen~\textit{et al.}~\cite{chen2019palvo} proposed PALVO, a visual odometry system based on a panoramic annular lens, which greatly increases the robustness to rapid motion and dynamic scenarios.
Seok~\textit{et al.}~\cite{seok2020rovins} proposed ROVINS, a robust omnidirectional visual-inertial odometry system utilizing visual feature constraints and IMU constraints with four large-FoV fisheye cameras.
Wang~\textit{et al.}~\cite{wang2022lf} proposed LF-VIO, a real-time visual-inertial-odometry framework designed for cameras with extremely large FoV.
For the second type of images, Sumikura~\textit{et al.}~\cite{sumikura2019openvslam} proposed OpenVSLAM, which is a VO framework supporting equirectangular images.
Huang~\textit{et al.}~\cite{huang2022360vo} proposed a novel direct visual odometry algorithm, called 360VO, which takes advantage of a single 360{\textdegree} camera for robust localization and mapping.

The aforementioned large-FoV SLAM frameworks only use point features as constraints without incorporating line features as additional constraints.
Moreover, the addition of line feature constraints has the potential to make the VIO system more robust and improve the accuracy of the localization system.

\begin{figure*}[!t]
%\vskip-1ex
	\centering
	\includegraphics[width=1.0\linewidth]{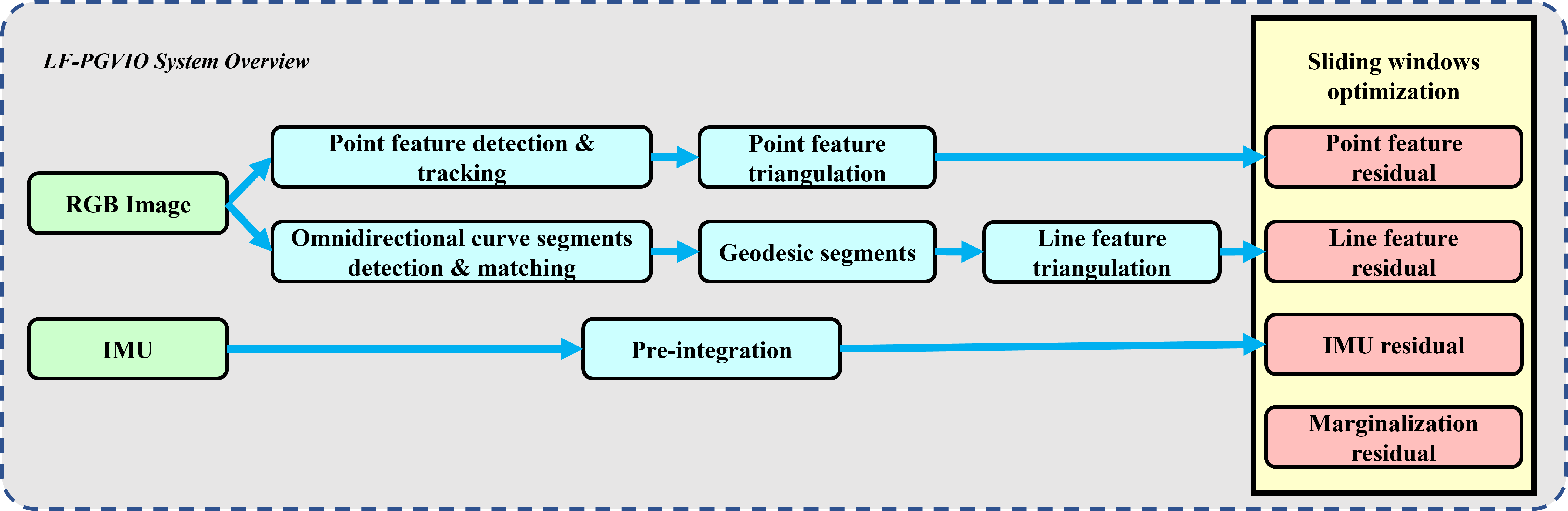}
	\vskip-1ex
	\caption{Overview of the proposed LF-PGVIO system. The sensor data includes RGB images and IMU. Our sliding window optimization approach consists of the following four parts: point feature residual, line feature residual, IMU residual, and marginalization residual.}
	\label{fig:overview}
	\vskip-3ex
\end{figure*}

\subsection{Point-line SLAM}
Compared with the SLAM systems using only point features as constraints, point-line SLAM systems using line features as constraints have the potential to reach higher accuracy and better robustness.
Pumarola~\textit{et al.}~\cite{pumarola2017pl} proposed PL-SLAM, a real-time monocular visual SLAM system that utilizes both points and lines to handle low-textured scenes.
{Zou~\textit{et al.}~\cite{zou2019structvio} proposed StructVIO, a novel visual-inertial odometry approach that adopts structural regularity in man-made environments.}
Fu~\textit{et al.}~\cite{fu2020pl} proposed PL-VINS, a real-time and high-efficiency optimization-based monocular visual-inertial SLAM method with point and line features.
Lim~\textit{et al.}~\cite{lim2022uv} proposed UV-SLAM, an unconstrained line-based SLAM method that uses vanishing points for structural mapping.
It addresses the limitations of existing line-based SLAM approaches by introducing UV-SLAM, which utilizes vanishing points for structural mapping without imposing constraints like the Manhattan world assumption.
Zhang~\textit{et al.}~\cite{zhang2023pli} proposed PLI-VIO, a real-time monocular visual-inertial odometry using point and line interrelated features.
They proposed an efficient line feature extraction algorithm based on EDlines and a line-to-point tracking method that exploits the interrelation between point and line features.
Feng~\textit{et al.}~\cite{feng2023improved} proposed a monocular visual-inertial odometry with point and line features using adaptive line feature extraction.
The method includes an adaptive image preprocessing algorithm to increase feature detection, a grouping and merging technique for line feature matching, and an initialization procedure utilizing least square estimation for acceleration bias. 

However, the above-mentioned works in point-line SLAM were based on the assumption that projecting a 3D straight line onto a plane results in a straight line using the image straight-line detection method. However, this approach fails when the field of view angle exceeds $90^\circ$.
To address this limitation, our proposed LF-PGVIO approach utilizes the projection of a 3-dimensional straight line onto a sphere, resulting in a great circle on the unit sphere. By employing curve detection, we ensure that the extracted curve is projected onto the unit sphere as a great circle, effectively overcoming the projection issues encountered with FoVs greater than $90^\circ$.

{Our system, unlike previous methods, supports large Field-of-View (FoV) images without requiring de-distortion operations. It directly extracts omnidirectional curve segments from the original image and adjusts line feature residuals to accommodate negative-plane FoV images and point-line SLAM systems with omnidirectional cameras.}

\section{Methodology}
In this paper, we propose LF-PGVIO, the overview of which is depicted in Fig~\ref{fig:overview}.
{LF-PGVIO mainly consists of two components front-end and back-end, the front-end consists of point feature and Omnidirectional Curve Segments (OCS) detection and matching, and the back-end solving the VIO optimization problem with constraint terms including point and line feature residual terms, IMU pre-integration residual terms, and marginalization residual terms.}

In Sec.~\ref{sec:geodesic_segments}, we introduce the theory of geodesic segment correlation, which provides a theoretical basis for omnidirectional curve segment extraction methods in the following images.
In Sec.~\ref{sec:curve_segment_drawing}, we propose the method of omnidirectional curve segment extraction methods based on the corresponding omnidirectional camera model, which can directly obtain the projected curve segments of 3D straight lines on raw large-FoV images.
In Sec.~\ref{sec:curve_segment_matching}, we propose the method of omnidirectional curve segment matching to provide observation data for line feature residual.
In Sec.~\ref{sec:tightly_coupled_lfpgvio},
we describe our proposed LF-PGVIO framework where we utilize the sliding windows optimization to solve the problem. In this framework, our introduced line feature residual supports large-FoV cameras, even with a negative imaging plane.

\subsection{Geodesic Segments} \label{sec:geodesic_segments}
First, we introduce the relationship between the geodesic segment, image segment, and 3D straight line segment, the characterization method and fitting of the geodesic segment, and the distance from the pixel to the omnidirectional curve segment.
It provides the basis for the extraction of omnidirectional curve segments. Specifically, $[x,y]^T$ is a point in the image space, and $[\alpha,\beta,\gamma]^T$ is a point on the unit sphere.

As a pixel $[x_m,y_m]^T$ on images with negative half-plane large-FoV can not be appropriately represented using $[u_m,v_m,1]^T$~\cite{wang2022lf}, we use $[\alpha_m,\beta_m,\gamma_m]^T$ to represent the pixel in the image as follows:
\begin{equation}
\begin{bmatrix} \alpha_m \\\beta_m \\\gamma_m\end{bmatrix}={P_s=}\pi^{-1}_s\left( \begin{bmatrix} x_m\\y_m\end{bmatrix}\right),
\end{equation}
\begin{equation}
\alpha_m^2+\beta_m^2+\gamma_m^2=1,
\end{equation}
where $\pi_s$ represents the mapping function from a point on the unit sphere to its corresponding pixel. The $\pi_s$ mapping function is determined by the camera model.

\begin{figure}[t!]
	\centering
	\includegraphics[width=0.9\linewidth]{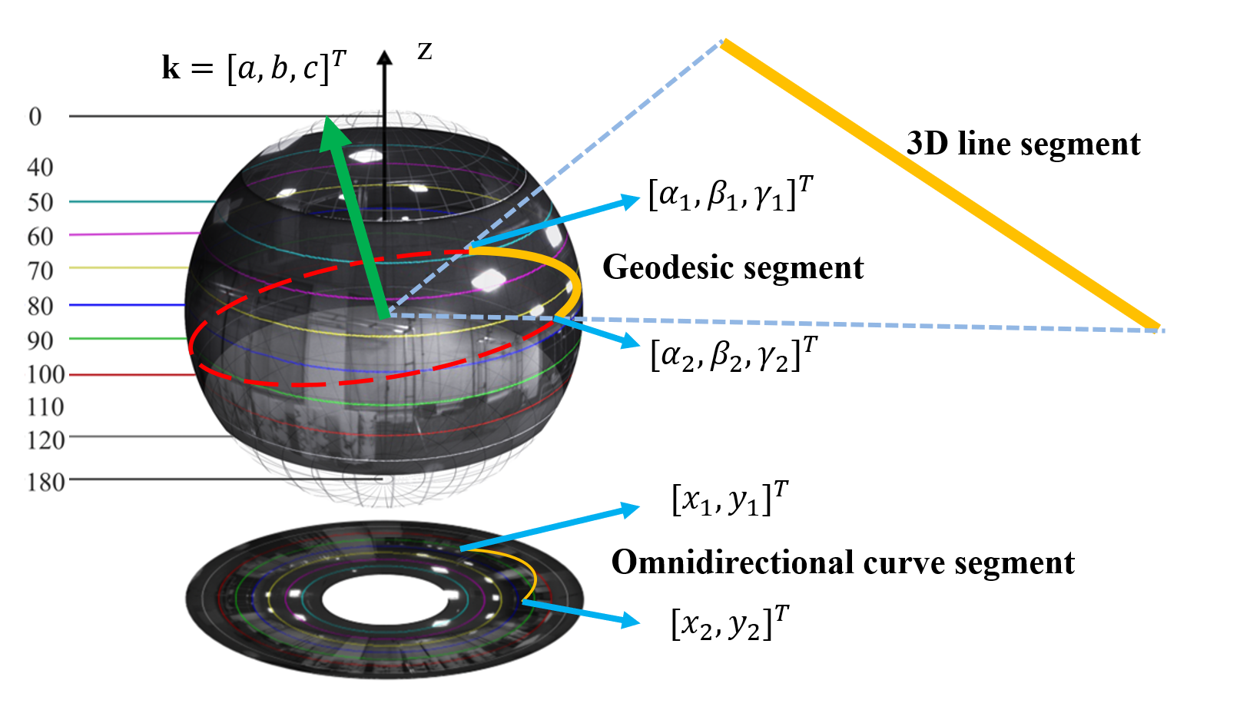}
	\vskip-1ex
	\caption{The projection from an orange 3D line onto a geodesic segment on a unit sphere, and a geodesic segment onto a curved segment on an image. The red dashed line is the great circle where the geodesic segment lies and the green vector represents one of the unit vectors that is perpendicular to the plane containing the great circle.}
	\label{fig:geodesic_seg}
	\vskip-3ex
\end{figure}

A line segment in $\mathbb{R}^3$ projected onto a unit sphere $\mathbb{S}^2$ is a geodesic segment in Fig.~\ref{fig:geodesic_seg}.
The two endpoints of the geodesic segment are $[\alpha_1,\beta_1,\gamma_2]^T$ and $[\alpha_2,\beta_2,\gamma_2]^T$.
The geodesic segment is the shortest segment between two points on a sphere. We denote by $\mathbf{k}$ the great circle on the sphere where the geodesic segment is located, where $\mathbf{k}$ is perpendicular to the great circle and its modulus is $1$.
The great circle is the circle determined by the two points on the sphere's surface that are farthest away from each other and whose diameter equals the diameter of the sphere.

\noindent\textbf{Great circle fitting:}
Given pixels set $I=\{[x_i,y_i]^T|(i=1\dots n)\}$ in an image, we can use the camera model to get its points set $S=\{[\alpha_i,\beta_i,\gamma_i]^T|(i=1\dots n)\}$ on the sphere. 
{This problem can be formulated as the following optimization problem:}
\begin{equation}
\begin{aligned}
&{\min\sum_{i=1\dots n}{(\alpha_ia+\beta_ib+\gamma_ic)^2}}\\
&{st. ||\mathbf{k}||_2=1.}
\label{eq:fitting}
\end{aligned}
\end{equation}
We use the Singular Value Decomposition (SVD) method to obtain one of the solutions for $k$ and the method of great circle fitting will be used for curve segment detection.

\begin{figure}[t!]
	\centering
	\includegraphics[width=0.5\linewidth]{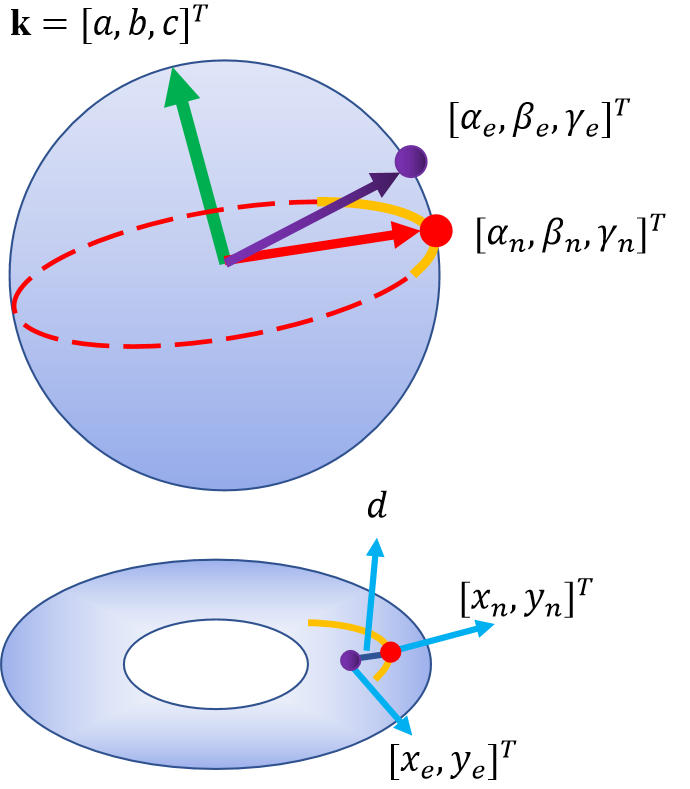}
	\vskip-2ex
	\caption{The distance between a purple pixel and an orange curve segment in the image. The purple and red points in the image correspond to their counterparts on the unit sphere, which are also purple and red respectively. The orange curve segment on the image corresponds to the orange geodesic segment on the unit sphere. This orange geodesic segment lies on the great circle represented by the red dashed line.  
 }
	\label{fig:distance}
	\vskip-3ex
\end{figure}

\noindent\textbf{The distance from the pixel to omnidirectional curve segment:}
Given a pixel $[x_e,y_e]^T$ in the image, its corresponding point on the sphere is $[\alpha_e,\beta_e,\gamma_e]^T$ based on the camera model.
The nearest point of $[\alpha_e,\beta_e,\gamma_e]^T$ on the sphere to the great circle in Fig.~\ref{fig:distance} is $[\alpha_n,\beta_n,\gamma_n]^T$, which can be obtained via the following equation:
\begin{equation} 
\begin{bmatrix} \alpha_n \\\beta_n \\\gamma_n\end{bmatrix}=\textbf{k}\times f_n(\begin{bmatrix} \alpha_e \\\beta_e \\\gamma_e\end{bmatrix}\times\textbf{k}),
\end{equation}
where $\textbf{k}$ denotes the vector of the perpendicular great circle and $f_n(\cdot)$ denotes the function that normalizes a vector 
$\mathbf{x}$ by dividing it by its Euclidean norm $||\mathbf{x}||_2$.
The distance $d$ from the pixel to the curve segment can be obtained as follows:
\begin{equation}
    d=f_d( \begin{bmatrix} x_e \\y_e\end{bmatrix},\mathbf{k})=||\pi_s( \begin{bmatrix} \alpha_n \\\beta_n \\\gamma_n\end{bmatrix})-\begin{bmatrix} x_e \\y_e\end{bmatrix}||_2,
    \label{eq:d}
\end{equation}
where $f_d(\cdot)$ is the function that computes the distance from the pixel to the omnidirectional curve segment and $\pi_s$ is the projection function from the unit sphere to the image plane. 

\subsection{Omnidirectional Curve Segment Detection (OCSD)} \label{sec:curve_segment_drawing}
Our omnidirectional curve segment detection algorithm has the following main steps:
1) Image pre-processing;
2) Computation of gradient magnitude and orientation;
3) Extraction of anchor pixels, which are the local maxima in the gradient magnitude;
4) Connect the anchors and determine whether the pixels are projected on the same great circle in the sphere based on the corresponding omnidirectional camera model.

For the image pre-processing step, we first use the histogram equalization algorithm and suppress the noise by using a Gaussian convolution kernel.
For the computation of gradient magnitude and orientation, we begin by applying the Sobel operator to calculate the horizontal gradient $(Gx)$ and vertical gradient $(Gy)$.
The L1 norm $(|Gx|+|Gy|)$ is then used to obtain the gradient magnitude. 

For the extraction of anchor pixels, the image pixels with a computed gradient magnitude greater than $0$ are scanned to check if they are local maxima in the gradient magnitude along the quantized direction of the gradient.
For a pixel with a vertical edge, its gradient orientation $O(x, y)$ indicates that it is an anchor if $G[x, y] - G[x-1, y] \geq T_{anchor}$ and $G[x, y] - G[x+1, y] \geq T_{anchor}$.
Similarly, if the gradient orientation indicates a horizontal edge, the pixel is an anchor if $G[x, y] - G[x, y-1] \geq T_{anchor}$ and $G[x, y] - G[x, y+1] \geq T_{anchor}$.

\begin{figure}[t!]
	\centering
	\includegraphics[width=1.0\linewidth]{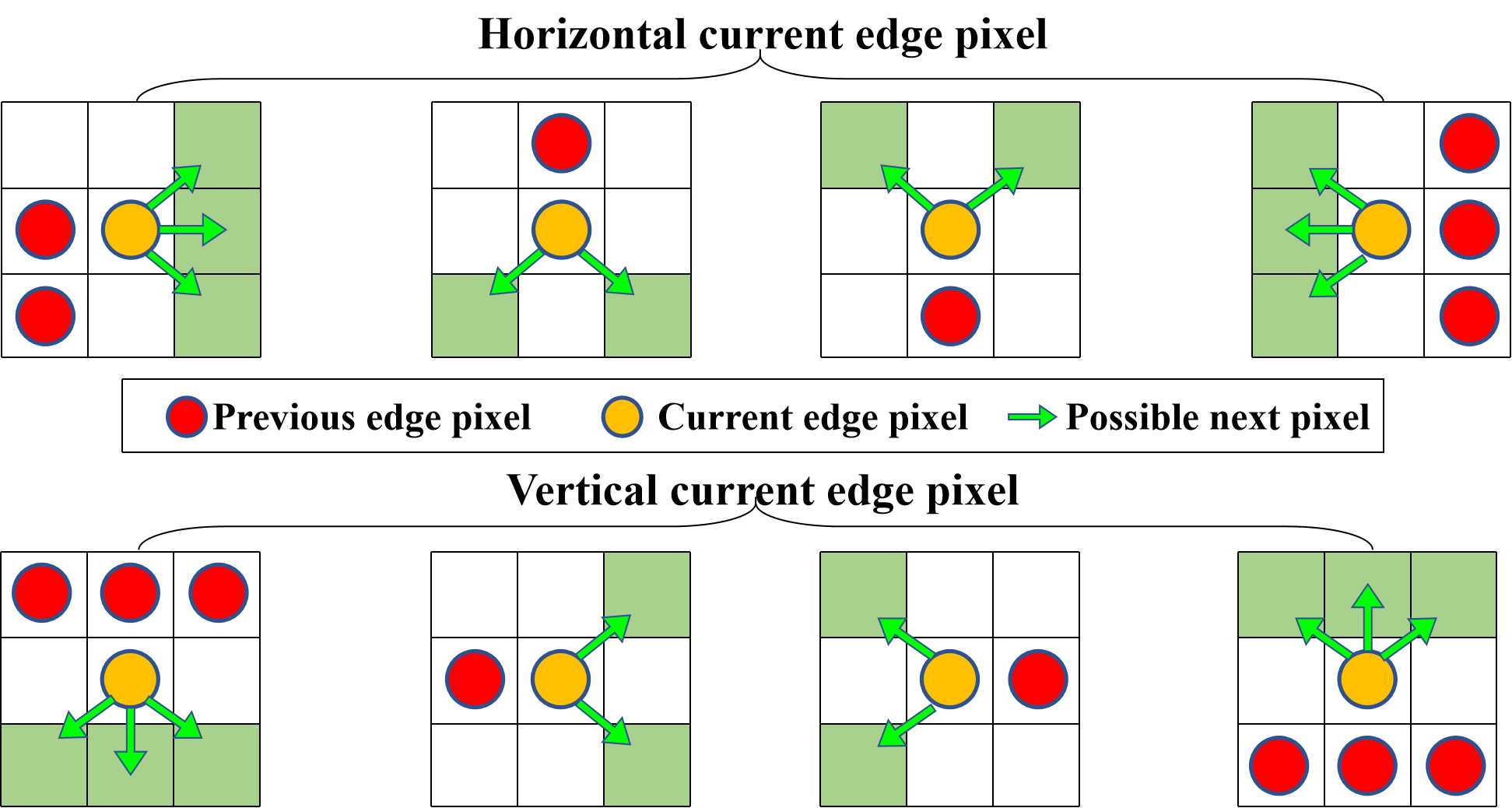}
	\vskip-1ex
	\caption{The pixel growth strategy of the EED algorithm. 
    The first row indicates that the current pixel is horizontal, and the second row indicates that the current pixel is vertical. Green arrows and squares indicate the possible direction and position of the next pixel.}
	\label{fig:EED}
	\vskip-3ex
\end{figure}

For connecting the anchors, starting from an initial anchor pixel, the algorithm only traverses along a chain of edge pixels, evaluating three neighboring pixels at each step.
Our OCSD procedure simultaneously performs omnidirectional curve edge detection and great circle fitting.
We consider the previous and current pixels during the omnidirectional curve edge detection process.
We follow the same approach as the EED algorithm~\cite{suarez2022elsed} and explore the feasible next pixels during the walking process according to the rule depicted in Fig.~\ref{fig:EED}. 
We use the pixel growth strategy to reduce the number of searches for adjacent pixels to reduce computational burdens.

When detecting omnidirectional curve edges, we use the distance from the pixel to the omnidirectional curve in Eq.~(\ref{eq:d}) to evaluate whether the point is added to the sequence. When a pixel point is added to the sequence, the current great circle parameters $[a,b,c]^T$ are re-estimated using the great circle fitting algorithm in Eq.~(\ref{eq:fitting}), which is described in Algorithm~\ref{alg:ECSED}.

Using the presented algorithm, we obtain an omnidirectional curve segment on the image, which is represented by the two endpoints of the omnidirectional curve segment. The extracted omnidirectional curve segments are prepared for descriptor matching in Sec.~\ref{sec:curve_segment_matching}.

\subsection{Omnidirectional Curve Segment Matching} \label{sec:curve_segment_matching}
We divide the omnidirectional curve segments into $n$ parts according to the corresponding major circle radians, and each sub-curve segment corresponds to $m^\circ$ on the unit sphere.
Each sub-curve segment is treated as a sub-linear segment to extract a Line Binary Descriptor (LBD) $\mathbf{D_l}$ according to \cite{zhang2013efficient}, and then these sub-linear segments are recombined into a new descriptor (RLBD) $\mathbf{D_c}$.
The recombined method is the same as LBD and the RLBD is used for matching.
The extracted omnidirectional curve segments are matched using the Hamming distance of the descriptor as a threshold condition for descriptor matching. The matched omnidirectional curve segments are utilized to construct line feature residuals.

\begin{algorithm}[t]
        \small
	\caption{Omnidirectional Curve Edge Detection Algorithm}
	\label{alg:ECSED}
	\KwIn{Anchor pixel: $a$. Anchor gradient direction: $d_0$}
	\KwOut{$S$: Set of omnidirectional curve Segments, $P$: Set of omnidirectional curve edge pixels}
	\Begin
	{
	    Initialization: $P \leftarrow \{a\}$;$S \leftarrow\emptyset$\;
            $\mathcal{D}_{stack} \leftarrow\emptyset$;$\mathcal{D}_{stack}$.push$([a,d_0])$\;
            \While{$\mathcal{D}_{stack}\neq\emptyset$}
            {
                $segmentFound\leftarrow$ $\textbf{false}$\;
                $nOutliers\leftarrow 0$\;
                $c,d$$\leftarrow$ $\mathcal{D}_{stack}$.pop()\;
                $p\leftarrow$previousPixel$(c, d)$\;
                \While{ $G[c]\neq0$ $\land$\ nOutliers $\leq T_{outliers}$}
                {
                    $c, p\leftarrow$  drawNextPx$(c, p)$\;
                    $P\leftarrow P\cup\{c\}$\;
                    \If{segmentFound}
                    {
                        $s, nOuliers\leftarrow$  addPxToGeodesicSegment$(s, c, nOutliers)$\;
                    }
                    \Else
                    {
                        $s, segmentFound\leftarrow$ fitGreatCircle$(P)$\;
                        $ S \leftarrow S \cup \{s\}$\;
                    }                
                }   
            }       
	}
\end{algorithm}

\subsection{Tightly-coupled LF-PGVIO} \label{sec:tightly_coupled_lfpgvio}

In this part, we describe the tightly-coupled back-end optimization of LF-PGVIO using sliding windows. By jointly processing and optimizing these measurements in a synchronized manner, LF-PGVIO utilizes measurements from two sensors to achieve highly accurate and robust pose estimation.
To integrate line features into a monocular large-FoV visual-inertial odometry system, the line features are extracted using the OCSD algorithm and matched using the RLBD.
The Plücker coordinate is an intuitive and elegant way to represent 3D lines~\cite{bartoli2005structure}, where a line is represented as $\mathbf{L}(\mathbf{n}, \mathbf{d})$ $\in$ $\mathbb{R}^{6}$, with $\mathbf{n}$ and $\mathbf{d}$ representing the normal and direction vectors, respectively.
The Pl\"ucker coordinate is utilized in the triangulation and re-projection process for line features, but it presents an over-parameterization problem in the optimization process of VIO since the Plücker coordinate represents 6 Degrees of Freedom (DoF) for a 3D line, while it has only 4-DoF.
To solve this problem, an orthonormal representation is employed to represent 3D lines with only 4-DoF.
{The orthonormal representation uses a rotation matrix, $\boldsymbol{\psi}$,} in Euler angles as the line's rotational component with respect to the camera coordinate system, and a scalar parameter, $\phi$, to represent the minimal distance from the omnidirectional camera center to the line.
Hence, the orthonormal representation can be expressed as follows:
\begin{equation}
    \mathbf{o} = [{\boldsymbol{\psi}},\phi]. 
\end{equation}
In LF-PGVIO, the state vector variables are shown as follows:
\begin{equation}
\begin{aligned}
\label{eq:var}
    \chi =& [\mathbf{x}_1,\dots,\mathbf{x}_{N},{\mathbf{t}_{c^0}^B,\dots,\mathbf{t}_{c^N}^B},\\
            & \lambda_{1},\dots,\lambda_{K},\\
            & \mathbf{o}_{1},\dots,\mathbf{o}_{J},]\\
     \mathbf{x}_n  =&[\mathbf{P}_{B_n}^W,\mathbf{V}_{B_n}^W,\mathbf{q}_{B_n}^W,\mathbf{b}_a,\mathbf{b}_g],n\in[0,N],\\
     \mathbf{t}_{c^i}^B = &[{\mathbf{P}_{c^i}^B},\mathbf{q}_{c^i}^B],\\
     \mathbf{o}_j =&[{\boldsymbol{\psi}_j},\phi_j],j\in[0,J].
\end{aligned}
\end{equation}
where $\mathbf{x}_n$ represents the state of the body in the $n$-th sliding window.
It includes the position $\mathbf{P}_{B_n}^W$, velocity $\mathbf{V}_{B_n}^W$, orientation quaternion $\mathbf{q}_{B_n}^W$, and biases of the accelerometer and gyroscope, denoted as $\mathbf{b}_a$ and $\mathbf{b}_g$, respectively.
$\mathbf{t}_{c^i}^B$ represents the transformation from the camera coordinate system $c^i$ to the body coordinate system.
It includes the position $\mathbf{P}_{c^i}^B$ and orientation quaternion $\mathbf{q}_{c^i}^B$. $\lambda_k$ represents the inverse distance of the $k$-th feature point from its first observation to the unit sphere\cite{wang2022lf}.
$\mathbf{o}_j$ represents the orthonormal representation of the $j$-th 3D straight line.
It includes the orientation angles, {which are denoted as $\boldsymbol{\psi}_j$ and $\phi_j$.}
Here, $N$ represents the total number of sliding windows utilized in the algorithm, whereas $K$ and $J$ denote the number of feature points and 3D straight lines, respectively.

By utilizing the states defined in Eq.~(\ref{eq:var}), the overall objective for optimization in the LF-PGVIO system can be expressed as follows:
\begin{equation}
\label{eq:optimization}
\begin{aligned}
&\min\limits_{\chi}
\Bigg\{
||\mathbf{r}_p-\mathbf{H}_p\chi||^2
+\sum_{k\in\mathbf{B}}||{\mathbf{r}_\mathbf{B}\left(\hat{\mathbf{z}}_{b_{k+1}}^{b_k},\chi\right)||^2_{\mathbf{P}^{b_k}_{b_{k+1}}}}\\
+&\sum_{\left(i,j\right)\in\mathbf{C}}\rho_p||\mathbf{r}_\mathbf{C}\left(\hat{\mathbf{z}}_{p}^{c_j},\chi\right)||^2_{\mathbf{P}^{c_j}_{p}}+\sum_{\left(i,j\right)\in\mathbf{L}}\rho_l||\mathbf{r}_\mathbf{L}\left(\hat{\mathbf{z}}_{l}^{c_j},\chi\right)||^2_{\mathbf{P}^{c_j}_{l}}
\Bigg\},
\end{aligned}
\end{equation}
where $\chi$ is the optimization variable. $\mathbf{r}_p$, $\mathbf{r}_\mathbf{B}$, $\mathbf{r}_\mathbf{C}$, $\mathbf{r}_\mathbf{L}$ represent marginalization,
IMU pre-integration, point, and line feature residuals, respectively.
$\mathbf{B}$ is the set of all pre-integrated IMU measurements in a sliding window. 
$\mathbf{C}$ and $\mathbf{L}$ are the sets of point and line measurements in observed frames.
$\mathbf{P}^{b_k}_{b_{k+1}}$, $\mathbf{P}^{c_j}_{p}$, $\mathbf{P}^{c_j}_{l}$ represent the covariance matrices of IMU, point, and line measurement.
To solve the nonlinear maximum posterior estimation problem in LF-PGVIO, the Ceres Solver algorithm~\cite{ceres} is utilized.
The Huber loss function $\rho(\cdot)$ is employed to reduce the impact of outliers, which leads to improved robustness of the overall visual-inertial odometry system.

\begin{figure}[t!]
	\centering
	\includegraphics[width=1.0\linewidth]{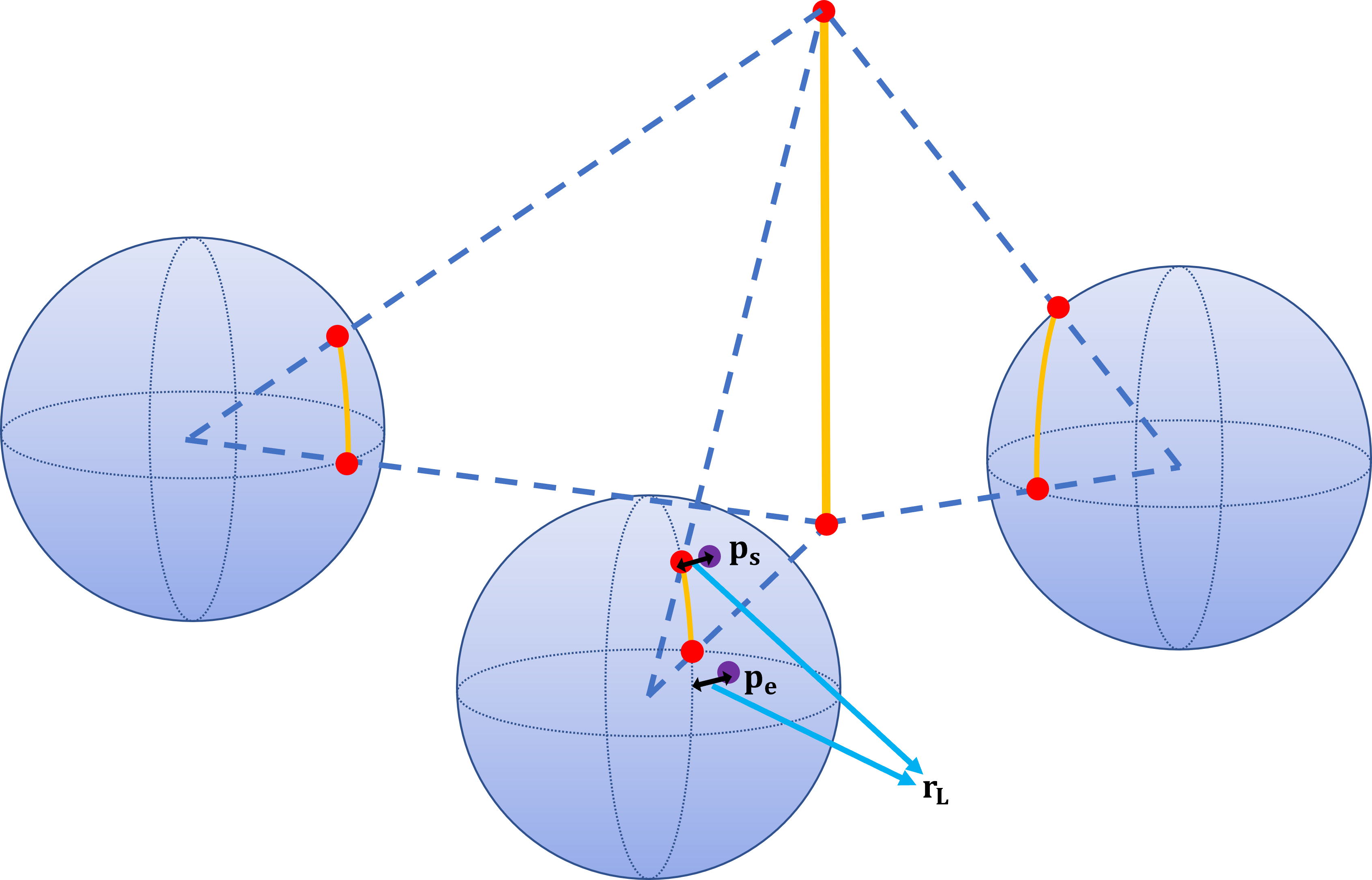}
	%\vskip-1ex
	\caption{Line feature residual. The orange 3D line is projected onto orange geodesic segments in different image frames. $\mathbf{p}_s$ and $\mathbf{p}_e$ are observed geodesic segments endpoints.}
	\label{fig:Line_res}
	\vskip-3ex
\end{figure}

\noindent\textbf{Line feature residual:} The line residual in previous works~\cite{fu2020pl,lim2022uv} does not support large-FoV cameras, especially those with a negative half-plane FoV. Therefore, instead of projecting the 3D space lines onto the $z=1$ plane, we first project them onto the unit spherical surface to address this issue. The residual of the line measurement model in the LF-PGVIO system is depicted in Fig.~\ref{fig:Line_res}. The re-projection error of the line measurement model is defined mathematically as follows:
\begin{equation}
    \mathbf{r}_\mathbf{L}=\begin{bmatrix}
        d(\mathbf{p}_s,\mathbf{n}^c)\\d(\mathbf{p}_e,\mathbf{n}^c)
    \end{bmatrix}
\end{equation}
where
\begin{equation}
\begin{aligned}
    &d(\mathbf{p},\mathbf{n}^c)=\frac{||\mathbf{p}^T\mathbf{n}^c||}{||\mathbf{n}^c||_2}\\
    &\mathbf{p}_s=[\alpha_s,\beta_s,\gamma_s]^T,\mathbf{p}_e=[\alpha_e,\beta_e,\gamma_e]^T.
\end{aligned}
\end{equation}
$\mathbf{r}_\mathbf{L}$ represents the line feature residual, and $d$ is the distance between the endpoints of the observed omnidirectional curve segment and its reprojected geodesic segment. The normal vector of the plane containing the reprojected geodesic segment is $\mathbf{n}^c$ and the endpoints of the observed omnidirectional curve segment on the unit sphere are denoted as $\mathbf{p}_s$ and $\mathbf{p}_e$.

\section{Experiments}
\begin{table*}[t!]
 \setlength{\tabcolsep}{2.6pt}
 \caption{Comparison of curve detection methods on the SUN360 indoor dataset~\cite{xiao2012recognizing}.}
 \label{tab:sun360}
  % \vskip-1ex
 \renewcommand\arraystretch{1.4}{\setlength{\tabcolsep}{0.8mm}{
\begin{tabular}{ccccccccccccccccccc}

\toprule 
     & \multicolumn{6}{c}{Fisheye image}                                                                           & \multicolumn{6}{c}{Panoramic annular images}                                                               & \multicolumn{6}{c}{Panoramic images}                                                                        \\ \cline{2-19} 
     & \multicolumn{2}{c}{$d_{orth}$} & \multicolumn{2}{c}{$d_s$} & \multirow{2}{*}{time$\downarrow$} & \multirow{2}{*}{\begin{tabular}[c]{@{}c@{}}\#lines\\ /image\end{tabular}} & \multicolumn{2}{c}{$d_{orth}$} & \multicolumn{2}{c}{$d_s$} & \multirow{2}{*}{time$\downarrow$} & \multirow{2}{*}{\begin{tabular}[c]{@{}c@{}}\#lines\\ /image\end{tabular}} & \multicolumn{2}{c}{$d_{orth}$} & \multicolumn{2}{c}{$d_s$} & \multirow{2}{*}{time$\downarrow$} & \multirow{2}{*}{\begin{tabular}[c]{@{}c@{}}\#lines\\ /image\end{tabular}} \\ \cline{2-5} \cline{8-11} \cline{14-17}
     & Rep-5$\uparrow$      & LE-5$\downarrow$       & Rep-5$\uparrow$         & LE-5$\downarrow$        &                       &                                & Rep-5$\uparrow$       & LE-5$\downarrow$       & Rep-5$\uparrow$         & LE-5$\downarrow$        &                       &                              & Rep-5$\uparrow$       & LE-5$\downarrow$       & Rep-5$\uparrow$         & LE-5$\downarrow$        &                       &                                \\ \hline
     \midrule

    Ours  & \textbf{0.531}&   \textbf{0.819}         &   \textbf{0.208}    &   \textbf{1.589}           &      \textbf{0.031}                             & 105.7    &   \textbf{0.560}        &   \textbf{1.472}        &      \textbf{0.253}        &     \textbf{1.637}       &           \textbf{0.075}            &     200.2                            &    \textbf{0.612}        &     \textbf{1.226}      &    \textbf{0.315}          &      \textbf{1.511}      &        \textbf{0.059}    &    218.2                  \\     

{RHT~\cite{torii2007randomized}} &     {0.223}       &  {1.124}        &  {0.023}          &  {2.183}    & {1.538}   &    {66.8}    &  {0.192}    &  {1.747}      &  {0.010}         & {1.374}         &  {1.753}             &  {90.0}         &   {0.304}            &   {1.378}        &  {0.047}        &  {2.341}  &   {1.810}   &       {120.1}      \\

ULSD~\cite{li2021ulsd} &      0.127      &   1.017        &      0.122        &     1.622        &        0.261              &         54.7          &      0.069      &     1.488      &     0.074         &     1.722       &               0.375        &                 38.6      &           0.176        &      1.248      &     0.187       &    1.527  & 0.443    &    79.0                            \\

                  	\bottomrule
\end{tabular}}}
% \vskip-1ex
\end{table*}

\begin{table*}[t]
 \setlength{\tabcolsep}{2.6pt}
 \caption{Comparison of curve detection methods on the CVRG-Pano outdoor dataset~\cite{orhan2022semantic}.}
 \label{tab:cvrg}
 % \vskip-1ex
 \renewcommand\arraystretch{1.4}{\setlength{\tabcolsep}{0.8mm}{
\begin{tabular}{ccccccccccccccccccc}

\toprule 
     & \multicolumn{6}{c}{Fisheye image}                                                                           & \multicolumn{6}{c}{Panoramic annular images}                                                                & \multicolumn{6}{c}{Panoramic images}                                                                        \\ \cline{2-19} 
     & \multicolumn{2}{c}{$d_{orth}$} & \multicolumn{2}{c}{$d_s$} & \multirow{2}{*}{time$\downarrow$} & \multirow{2}{*}{\begin{tabular}[c]{@{}c@{}}\#lines\\ /image\end{tabular}} & \multicolumn{2}{c}{$d_{orth}$} & \multicolumn{2}{c}{$d_s$} & \multirow{2}{*}{time$\downarrow$} & \multirow{2}{*}{\begin{tabular}[c]{@{}c@{}}\#lines\\ /image\end{tabular}} & \multicolumn{2}{c}{$d_{orth}$} & \multicolumn{2}{c}{$d_s$} & \multirow{2}{*}{time$\downarrow$} & \multirow{2}{*}{\begin{tabular}[c]{@{}c@{}}\#lines\\ /image\end{tabular}} \\ \cline{2-5} \cline{8-11} \cline{14-17}
     & Rep-5$\uparrow$      & LE-5$\downarrow$       & Rep-5$\uparrow$         & LE-5$\downarrow$        &                       &                                & Rep-5$\uparrow$       & LE-5$\downarrow$       & Rep-5$\uparrow$         & LE-5$\downarrow$        &                       &                              & Rep-5$\uparrow$       & LE-5$\downarrow$       & Rep-5$\uparrow$         & LE-5$\downarrow$        &                       &                                \\ \hline
     \midrule

Ours &   \textbf{0.480}         &    \textbf{0.790}       &   \textbf{0.263}           &    \textbf{1.552}        &       \textbf{0.035}               &       107.2                         &          \textbf{0.510}  &             \textbf{1.262}         &   \textbf{0.300}         &       \textbf{1.650}                &     \textbf{0.049}                         &   181.2         &   \textbf{0.602}        &     \textbf{1.323}         &   \textbf{0.316}         & \textbf{1.525}                      &  \textbf{0.056} &  212.2   \\

{RHT~\cite{torii2007randomized}} &  {0.308}          &          {1.115}        &    {0.034}       &   {2.247}                &       {1.503}          &  {79.8}   & {0.352}        & {1.433}       &   {0.018}        &  {2.404}        &   {1.775}            & {135.9}          &    {0.314}         &       {1.869}    &   {0.033}       &  {2.566}   &  {1.905}   &            {160.5}                    \\

ULSD~\cite{li2021ulsd} &   0.087         &    0.799       &                  0.114       &       1.690                &            0.261                    &     41.0       &   0.036        &      1.363        &     0.026       &           1.733            &             0.341                 &    10.0        &    0.004       &      1.471                &       0.012                &         1.653    &       0.241     &       15.2       \\
                 	\bottomrule
\end{tabular}}}
% \vskip-3ex
\end{table*}

\subsection{Curve Segments Detection Comparison}
To validate the effectiveness of our proposed OCSD algorithm, we conduct extensive experiments utilizing three distinct types of large-FoV images: fisheye, panoramic annular, and panoramic images, which correspond to the MEI model~\cite{mei2007single}, the camera model introduced by Scaramuzza~\textit{et al.}~\cite{scaramuzza2006toolbox}, and the equirectangular camera model, respectively.
These experiments are carried out using the SUN360 dataset~\cite{xiao2012recognizing} and CVRG-Pano~\cite{orhan2022semantic}, which collectively contain $72$ indoor test images and $76$ outdoor test images.
Following the SOLD2 evaluation tool~\cite{pautrat2021sold2}, we utilize the rotation matrix to generate pairs of matching images for testing.

\noindent\textbf{Curve segment distance metrics:}  
In order to accurately evaluate the precision of curve segment detection, it is necessary to establish a curve segment distance metric.
Our work employs distance metrics of two curve segments for evaluation, detailed as follows:

\underline{Orthogonal distance} ($d_{orth}$): 
The orthogonal distance of two curve segments $c_1$ and $c_2$ is defined as $d_{orth}$:
\begin{equation}
\begin{aligned}
     &d_{a1}(c_1,c_2)=\frac{f_d(e_1^1,\mathbf{k}_2)+f_d(e_1^2,\mathbf{k_2})}{2},\\
     &d_{a2}(c_1,c_2)=\frac{f_d(e_2^1,\mathbf{k}_1)+f_d(e_2^2,\mathbf{k_1})}{2},\\
     &d_{orth}=\frac{d_{a1}+d_{a2}}{2},
\end{aligned}     
\end{equation}
where $\mathbf{k}_1$ and $\mathbf{k}_2$ are the great circle parameters corresponding to the curve segments $c_1$ and $c_2$ respectively. $f_d(\cdot)$ denotes the function that computes the distance from the pixel point to the curve segment. $(e_1^1,e_1^2)$ and $(e_2^1,e_2^2)$ are the endpoints of $c_1$ and $c_2$ respectively. 
When searching for the closest curve segment using this distance metric, curve segments with an overlap below $0.5$ are disregarded. This criterion enables curve segments that pertain to the same 3D line but possess varying lengths to be viewed as proximal to each other. Such an approach can be advantageous in VIO or SLAM assignments.

\underline{Structural distance} ($d_s$):
\begin{equation}
\begin{aligned}
     d_s(c_1,c_2)=\min(&\frac{||e_1^1-e_2^1||_2+||e_1^2-e_2^2||_2}{2},\\
     &\frac{||e_1^1-e_2^2||_2+||e_1^2-e_2^1||_2}{2})
\end{aligned}
\end{equation}
The structural distance is mainly used to evaluate the consistency of endpoint detection of curve segments.
By employing these distance metrics of two different curve segments, we are able to provide a comprehensive evaluation of the precision of our curve segment detection algorithm.

\noindent\textbf{Curve segment detection metrics:}  
As we aim to extract reliable and consistent curve segments from panoramic images, evaluating our curve segment detection method on hand-labeled curve segments of the dataset is not suitable.
Therefore, we modify the assessment metrics originally proposed in SuperPoint~\cite{detone2018superpoint} to evaluate the effectiveness of our curve segment detector. This is accomplished by utilizing pairs of rotated large-FoV images. 

\underline{Repeatability:} This metric measures the frequency with which a curve segment can be re-detected across different views.
It computes the average percentage of curve segments which appear in the first image and have a corresponding match when projected onto the second image.
Two curve segments are considered a match when their distance is below a specified threshold $\epsilon$. This metric is calculated symmetrically across the two images and averaged.

\underline{Localization error:} The localization error with tolerance $\epsilon$ refers to the average distance between a curve segment and its re-detection in the second image, considering only the matched curve segments.

\begin{figure*}[t]
	\centering
	\includegraphics[width=1.0\linewidth]{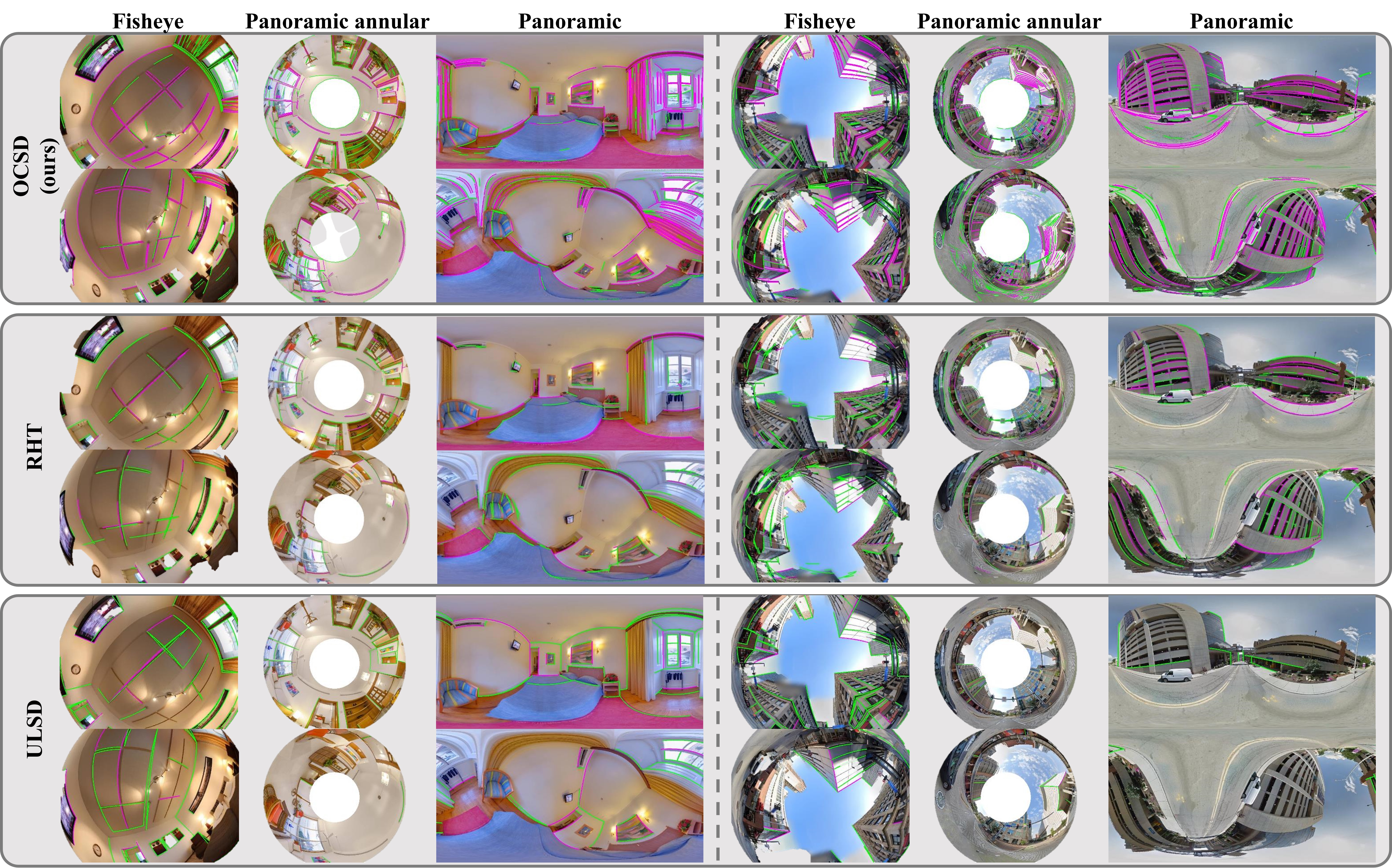}
	%\vskip-1ex
        \caption{
        {Comparison of curve detection visualizations between our proposed method OCSD, RHT, and ULSD on two datasets: SUN360 (on the left) and CVRG-PANO (on the right).}
        The \purple{purple} curve segments in the image denote the matched pairs with $d_{orth}$ less than $5$ pixels and overlap greater than $0.5$. The \green{green} curve segments represent unmatched pairs.}
        \vskip-3ex
	\label{fig:CSED_benchmark}
\vskip-1ex
\end{figure*}

\noindent\textbf{Comparison on the public datasets:} 
Targeting omnidirectional data, we compare our proposed OCSD method against RHT~\cite{torii2007randomized} and the state-of-the-art ULSD~\cite{li2021ulsd}.
The results are compared on the SUN360 indoor dataset~\cite{xiao2012recognizing} and the CVRG-Pano outdoor dataset~\cite{orhan2022semantic}, respectively, as shown in Table~\ref{tab:sun360} and Table~\ref{tab:cvrg}. 
Our OCSD method is implemented with a minimum curve segment length of $30$ pixels.
ULSD has been trained on the SUN360 dataset, thus its pre-trained parameters are directly utilized.
As the CVRG-Pano dataset does not contain annotated line features, we are unable to retrain ULSD on this dataset and reuse the weight of the SUN360 dataset.

To demonstrate the versatility and applicability of our algorithm to various large-FoV cameras, we create a dataset consisting of three types of images: fisheye images, panoramic annular images, and panoramic images. 
For the dataset, each type of image is subjected to an arbitrary rotation matrix to generate corresponding pairs based on different camera models. The image sizes are $720{\times}540$, $1280{\times}960$, $1024{\times}512$, respectively. For fisheye and panoramic annular images, the rotation angle is limited to $60^\circ$.
For panoramic images, there is no panoramic angle limit.
Our results indicate that our approach achieves superior performance in terms of repeatability and localization error on both datasets and the aforementioned three types of images when compared to RHT and ULSD. 

Additionally, we demonstrate the effectiveness of our metric, $d_{orth}$, through visualization in Fig.~\ref{fig:CSED_benchmark}. 
It can be seen that compared with RHT and ULSD, our method has more matched pairs with $d_{orth}$ less than $5$ pixels and overlap greater than $0.5$.
In Fig.~\ref{fig:CSED_benchmark}, the results on various large field-of-view images, including fisheye, panoramic annular, and panoramic images taken indoors or outdoors, demonstrate that our method achieves a higher matching rate compared to ULSD. This can be observed by the presence of more purple lines and fewer blue lines in the image. 
ULSD, trained solely on indoor datasets, exhibits a significant drop in performance when tested on unknown outdoor datasets. 
This indicates that the performance of ULSD is heavily reliant on the data distribution of the dataset. In contrast, our algorithm OCSD demonstrates superior matching effectiveness and better generalization compared to ULSD.
With robust curve feature extraction, the VIO system could reach better accuracy and reliability. Moreover, our OCSD algorithm consumes less time compared to RHT and OCSD in Table~\ref{tab:sun360} and Table~\ref{tab:cvrg}, so this curve detection method is more suitable as a VIO front-end to reduce the computational burden of VIO.

\begin{table*}[t!] 
 %\scriptsize
 %\footnotesize
 \setlength{\tabcolsep}{2.2pt}
	\centering
	\begin{threeparttable}
	\caption{Comparison of VIO methods on the TUM Room dataset.\protect\\$\mathbf{p}$: point constraint, $\mathbf{pg}$: point and geodesic segment constraints.}
	\label{tab:TUM_lf-pgvio}
    %\vskip-1ex
% \begin{tabular}{cccccccccccc}
\renewcommand\arraystretch{1.4}{\setlength{\tabcolsep}{6.0mm}{\begin{tabular}{cccccccc}
\toprule   
\multicolumn{2}{c}{\multirow{2}{*}{VIO-Method}} & \multicolumn{6}{c}{Sequences}                                                                                                                                          \\
\multicolumn{2}{c}{}                            & Room1           & Room2           & Room3           & Room4           & Room5           & Room6                      \\
\midrule
\midrule

\multirow{3}{*}{LF-PGVIO ($\mathbf{pg})$}      & RPEt (\%)         & \textbf{0.253} & \textbf{0.308} & \textbf{0.339} & \textbf{0.405} & \textbf{0.616} & \textbf{0.596} \\
                             & RPEr (degree/m)  & \textbf{0.036} & \textbf{0.028} & \textbf{0.049}          & 
                             
                            \textbf{0.057}          & \textbf{0.032}          & \textbf{0.078}      \\
                             & ATE (m)           & \textbf{0.078} & \textbf{0.096} & \textbf{0.133} & \textbf{0.045} & \textbf{0.206}          & \textbf{0.100}  \\
\midrule

\multirow{3}{*}{LF-VIO ($\mathbf{p}$)~\cite{wang2022lf}}      & RPEt (\%)         & 0.271 & 0.309 & 0.343 &  0.480 & 0.658 &  0.673  \\
                             & RPEr (degree/m)  & 0.040 & 0.032 & 0.043          & {0.064}         & 0.037         & 0.106          \\
                             & ATE (m)           & 0.094 & 0.115 & 0.162 & 0.061 & 0.236          & 0.120  \\

	\bottomrule
% 	\end{tabular}
	\end{tabular}}}
	\end{threeparttable} 
	%\vskip-3ex
\end{table*}

\begin{figure*}[t]
	\centering
	\subfigure[Top Trajectory on Room1]{
		\includegraphics[width=0.3\textwidth]{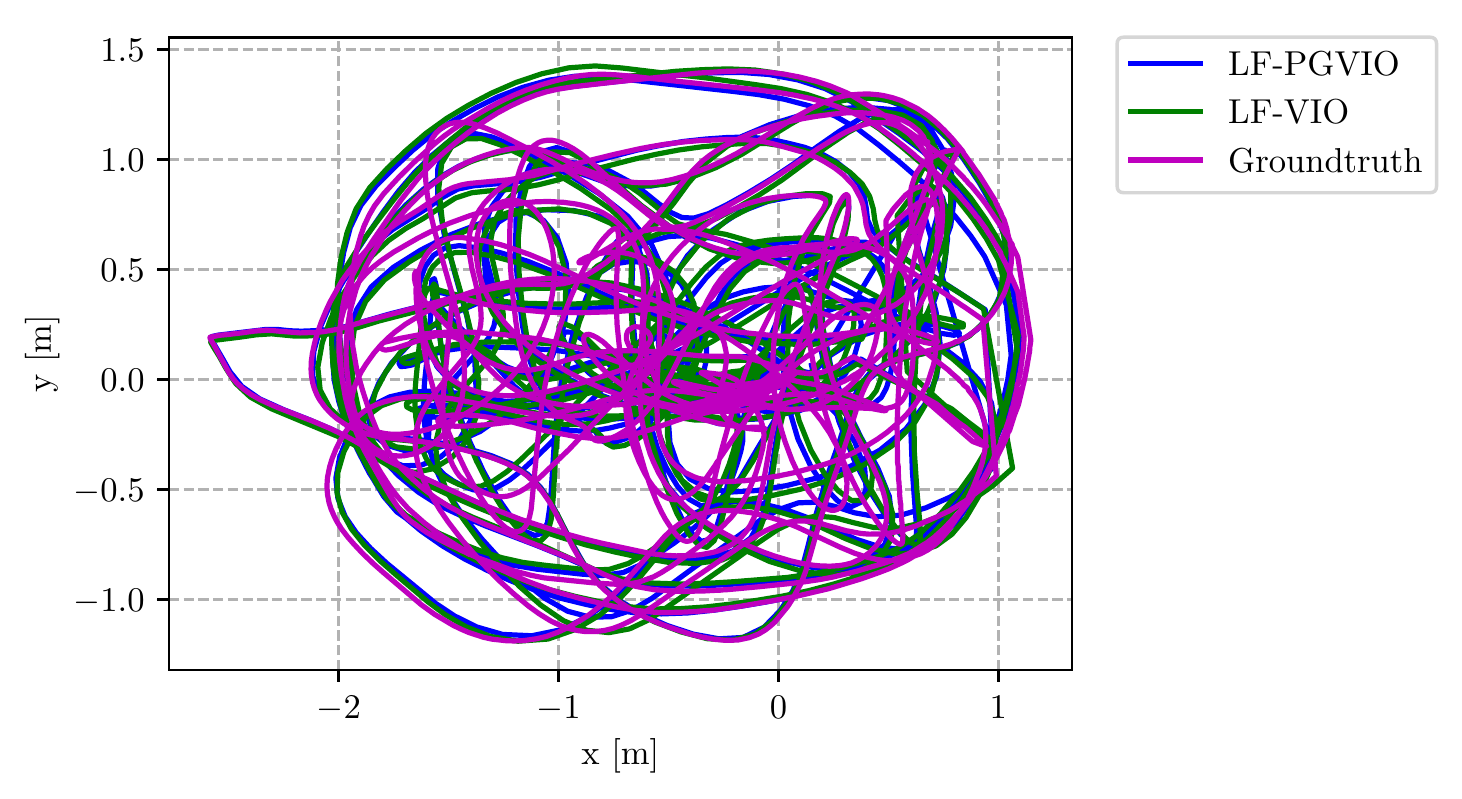}
  
	}
	\subfigure[Top Trajectory on Room4]{
		\includegraphics[width=0.3\textwidth]{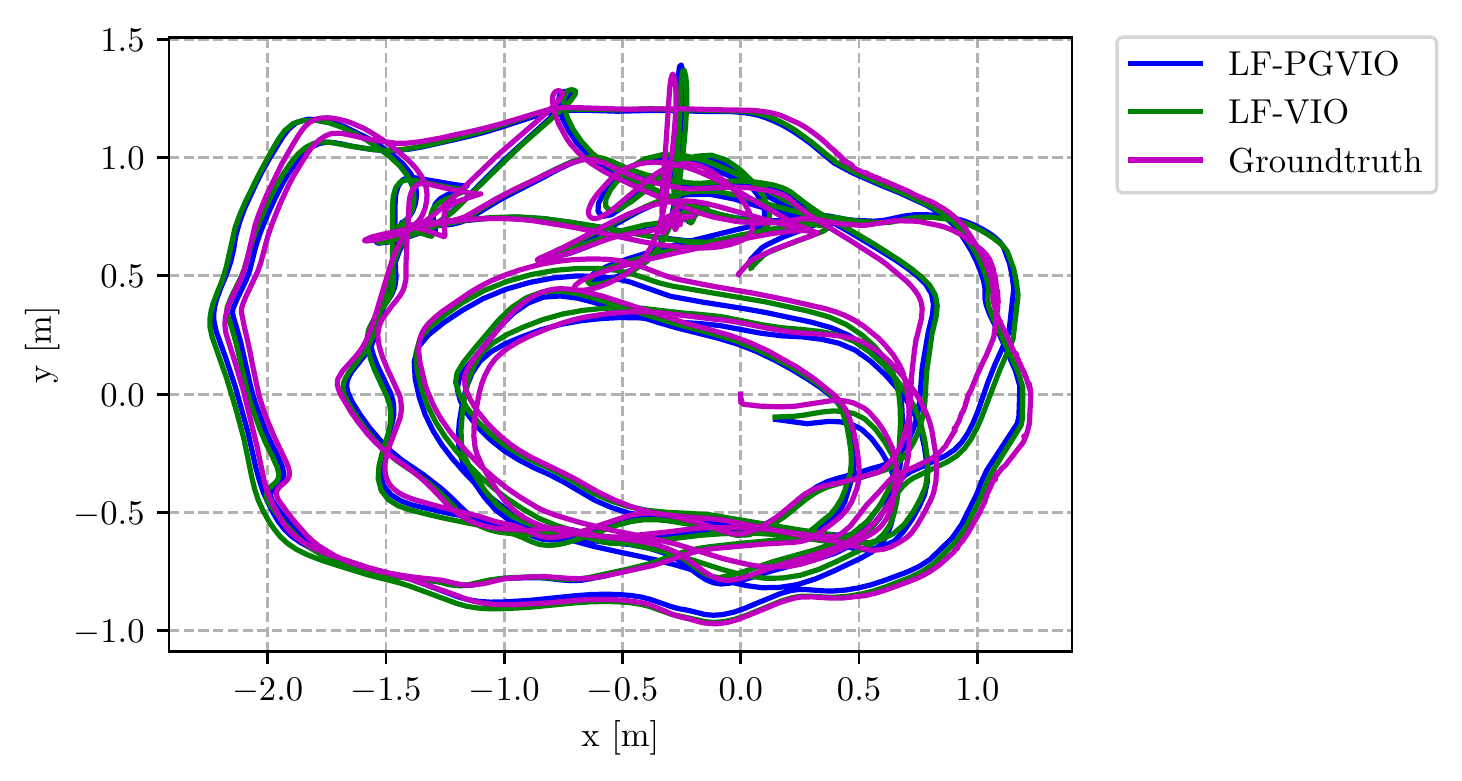}
		\label{fig:TUM_04_trajectory_top}
	}
 	\subfigure[Top Trajectory on Room6]{
		\includegraphics[width=0.3\textwidth]{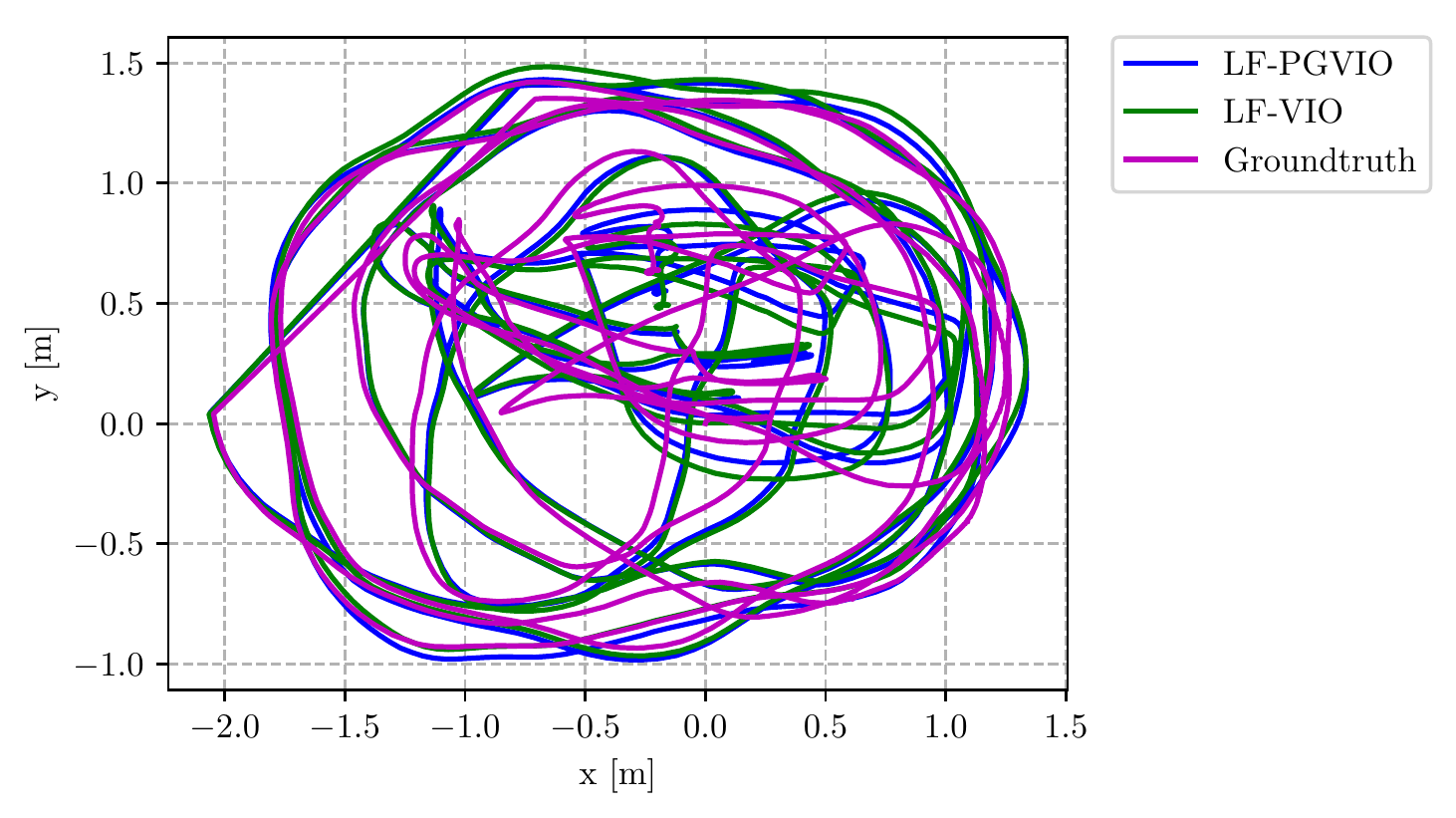}
		\label{fig:TUM_06_trajectory_top}
	}
	\subfigure[Translation and Rotation Error on Room1]{
		\includegraphics[width=0.3\textwidth]{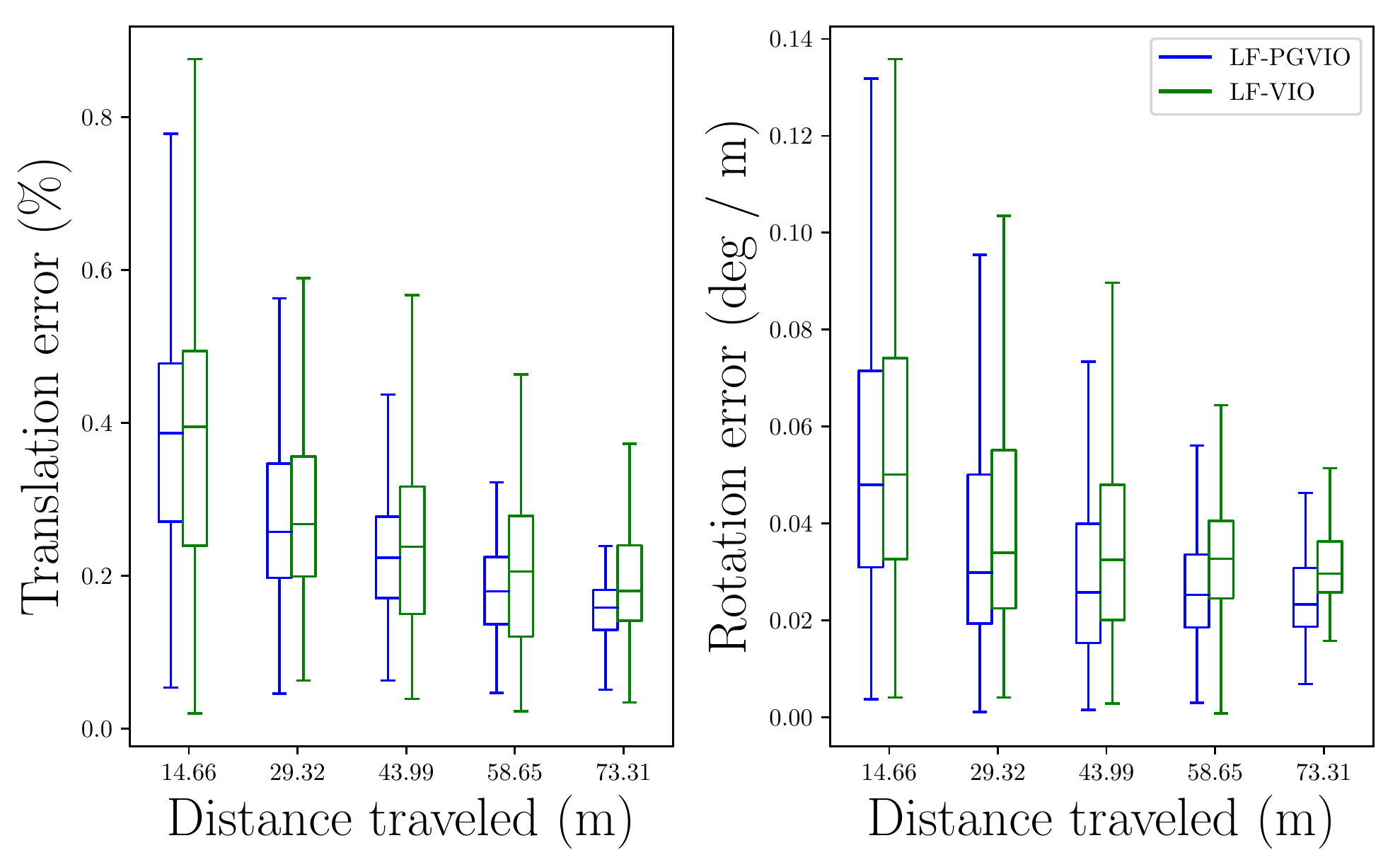}
		\label{fig:TUM_01_trans_rot_error}
	}
	\subfigure[Translation and Rotation Error on Room4]{
		\includegraphics[width=0.3\textwidth]{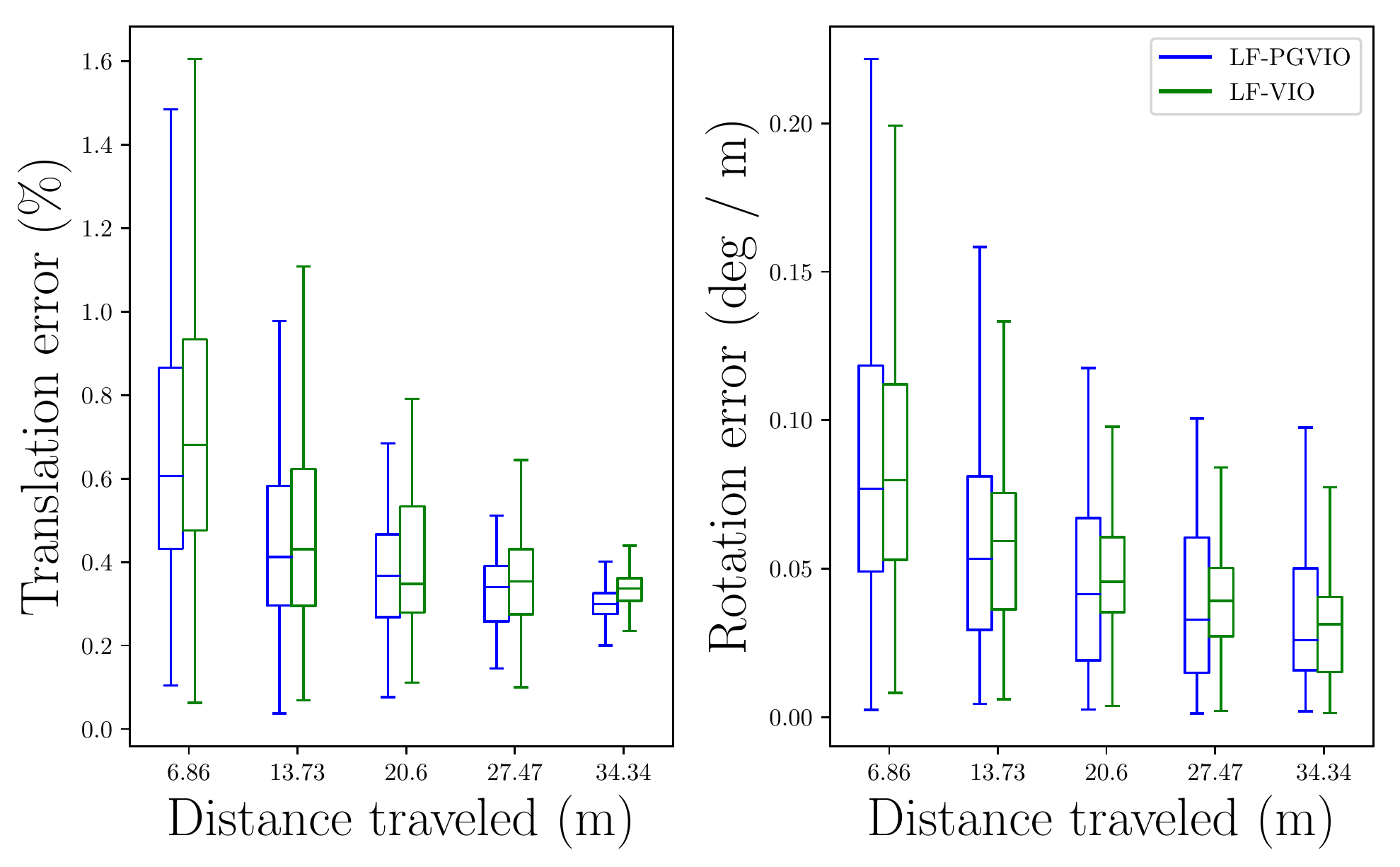}
		\label{fig:TUM_04_trans_rot_error}
	}
 	\subfigure[Translation and Rotation Error on Room6]{
		\includegraphics[width=0.3\textwidth]{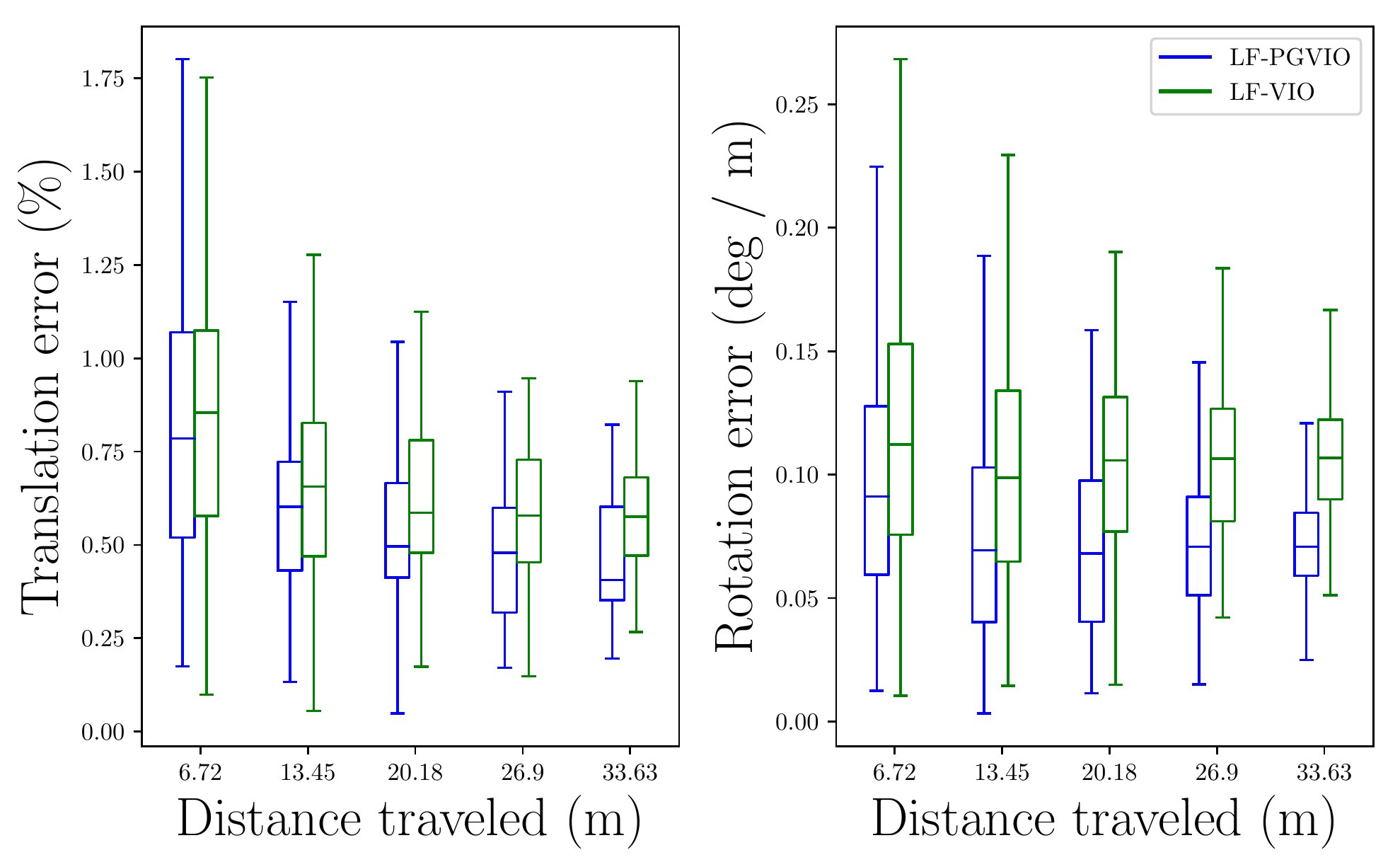}
		\label{fig:TUM_06_trans_rot_error}
	}
	%\vskip-1ex
	\caption{Examples of top trajectories and error analyses of different VIO systems on the TUM Room benchmark~\cite{schubert2018tum}.}
	\label{fig:TUM_lfvio-vins-svo}
	%\vskip-1ex
\end{figure*}

\begin{table*}[t!] 
 \setlength{\tabcolsep}{2.2pt}
	\centering
	\begin{threeparttable}
	\caption{Comparison of VIO methods on the PALVIO dataset~\cite{wang2022lf}.\protect\\$\mathbf{p}$: point constraint, {$\mathbf{pl}$: point and line segment constraints,} $\mathbf{pg}$: point and geodesic segment constrain.}
	\label{tab:lf-pgvio}
    % \vskip-1ex
\renewcommand\arraystretch{1.4}{\setlength{\tabcolsep}{1.8mm}{\begin{tabular}{cccccccccccccc}
\toprule   
\multicolumn{2}{c}{\multirow{2}{*}{VIO-Method}} & \multicolumn{10}{c}{Sequences}                                                                                                                                          \\
\multicolumn{2}{c}{}                            & ID01           & ID02           & ID03           & ID04           & ID05           & ID06           & ID07           & ID08           & ID09           & ID10  & {OD01}    & {OD02}   \\
\midrule
\midrule

\multirow{3}{*}{LF-PGVIO ($\mathbf{pg})$}      & RPEt (\%)         & \textbf{3.267} & \textbf{2.667} & \textbf{2.470} & \textbf{1.343} & \textbf{1.913} & \textbf{2.599} & \textbf{2.061} & 2.655          & \textbf{1.550}          & \textbf{3.539}  & {\textbf{4.446}}& {\textbf{4.738}}\\
                             & RPEr (degree/m)  & \textbf{0.313} & 0.409 & \textbf{0.227}          & {0.183}          & \textbf{0.252}          & 0.391          & 0.223          & {0.238} & 0.297 & \textbf{0.547} & {0.725}   &{\textbf{0.568}}       \\
                             & ATE (m)           & \textbf{0.332} & \textbf{0.147} & \textbf{0.265} & \textbf{0.143} & 0.182          & \textbf{0.089} & \textbf{0.220} & \textbf{0.181}          & \textbf{0.122} & \textbf{0.240}&{\textbf{0.263}} &{\textbf{0.288}} \\
\midrule

\multirow{3}{*}{LF-VIO ($\mathbf{p}$)\cite{wang2022lf}}      & RPEt (\%)         & {3.556} & {2.709} & {2.542} & {1.495} & {2.016} & 2.814 & {2.775} & 2.983          & 2.146          & {4.493} &{6.615} & {4.747} \\
                             & RPEr (degree/m)  & {0.328} & 0.599 & 0.292          & 0.227          & 0.328          & 0.397          & 0.315          & \textbf{0.202} & {0.322} & {0.567} &{0.506}  &{0.580}       \\
                             & ATE (m)           & {0.341} & {0.153} & {0.269} & {0.166} & 0.200          & {0.093} & {0.237} & 0.236          & {0.222} & {0.292}  &{0.362} &{0.292}\\
\midrule
\multirow{3}{*}{SVO2.0 ($\mathbf{p}$)~\cite{forster2014svo}}      & RPEt (\%)         & 6.531          & 6.995          & 2.710          & 1.928          & 2.354          & 3.409          & 3.718          & 2.811          & {2.012} & 14.147  & {34.799} &  { 26.408}    \\
                             & RPEr (degree/m)  & 0.401          & {0.378}         & {0.235} & \textbf{0.165}          & 0.296          & \textbf{0.320} & \textbf{0.187} & 0.291          & \textbf{0.230}          & 0.608 &{\textbf{0.301}} &{ 0.715}\\
                             & ATE (m)           & 0.761          & 0.380          & 0.366          & 0.174          & \textbf{0.148} & 0.124          & 0.428          & 0.236          & 0.292          & 1.122  &{2.674}  &{1.875}     \\
\midrule
\multirow{3}{*}{VINS-Mono ($\mathbf{p}$)~\cite{qin2018vins}}   & RPEt (\%)         & 5.446          & 3.542          & 2.767          & 2.189          & 2.553          & 2.993          & 2.941          & \textbf{2.405} & 2.933          & 4.494    & {21.181}&{$failed$}    \\
                             & RPEr (degree/m)  & 0.458         & 0.605          & 0.285          & 0.249          & {0.278} & 0.445          & 0.339          & 0.452          & 0.457          & {0.567}  &{0.539}  &{$failed$}     \\
                             & ATE (m)           & 0.870          & 0.214          & 0.310          & 0.217          & 0.263          & 0.104          & 0.299          & {0.194} & 0.378          & 0.557 &{1.171}&{$failed$}  \\ 
\midrule
\multirow{3}{*}{{PL-VINS ($\mathbf{pl}$)~\cite{fu2020pl}}}   & {RPEt (\%)}        &  {9.302}          &  {5.082}                 &  {5.572}         &{2.796}    & {3.898}        & {3.979}          &  {3.715}         &{4.586}  &   {3.154}        &   {12.880} &{30.221}  &{51.150}   \\
                             & {RPEr (degree/m)}  &  {0.530}        &   {0.894}        &   {0.468}        &   {0.342}        &{0.393}  &    {0.500}      &{0.377}          &   {0.710}        &    {0.435}      & {0.559} &{0.490} &{0.697}      \\
                             &{ATE (m)}           & {0.932} & {0.235} & {0.346} & {0.230} & {0.294} & {0.120} & {0.331} & {0.216} & {0.420} &  {0.601} &{2.843 }&{3.728}   \\ 
                            
\midrule
\multirow{3}{*}{{UV-SLAM ($\mathbf{pl}$)~\cite{lim2022uv}}}   & {RPEt (\%)}   &      {16.678}        &    {6.000}       &     {3.885}        &      {23.954}      &   {4.852}         &   {8.326}         &    {4.322}        &  {5.285} &   {6.105}         &   {32.234} &{ 22.279}   &{$failed$}   \\
                             & {RPEr (degree/m)}  &   {0.586}        &   {\textbf{0.344}}         &   {0.228}         &  {0.180}          &  {0.259} &     {0.411}      &    {0.246}       &   {0.373}         &      {0.288}     &     {0.624}&  {0.576}  &{$failed$}  \\
                             &{ATE (m)}           &   {2.616}       &   {0.572}       &{0.429}          &  {4.415}         &{0.561}      & {0.419} & {0.469} & {0.529} & {0.733} &  {2.678}  &{2.319}  &{$failed$}\\ 	

	\bottomrule
	\end{tabular}}}

	\end{threeparttable}
 
	%\vskip-3ex
\end{table*}

\subsection{VIO Comparison with the State-of-the-Art}

\noindent\textbf{Results on the TUM visual-inertial dataset}: 
To verify the validity of the line feature residual term, we conduct experiments on the public TUM visual-inertial odometry dataset~\cite{schubert2018tum}.
The IMU frequency of this data set is $200Hz$, the image resolution is $512{\times}512$ and the frequency is $20Hz$.
We use the monocular data from the dataset.
As the indoor dataset has rich line features and the motion capture device provides the ground truth at $125Hz$, we evaluate the VIO performance on the indoor dataset.
The fisheye FoV is $360^\circ{\times}(0^\circ{\sim}110^\circ)$.

For LF-VIO and LF-PGVIO, we both use default intrinsic and extrinsic camera parameters provided by the dataset and the same parameter setting except for line feature parts. {In the experiment on the TUM dataset, the maximum number of feature points of the image is set to $200$, and the sliding window size is $10$.}
In Table~\ref{tab:TUM_lf-pgvio} and Fig.~\ref{fig:TUM_lfvio-vins-svo}, it shows that LF-PGVIO yields more accurate results in Relative Pose Error in translation (RPEt), Relative Pose Error in rotation (RPEr), and Absolute Translation Error (ATE) in all Room sequences.

\begin{figure}[t]
	\centering
	\includegraphics[width=0.8\linewidth]{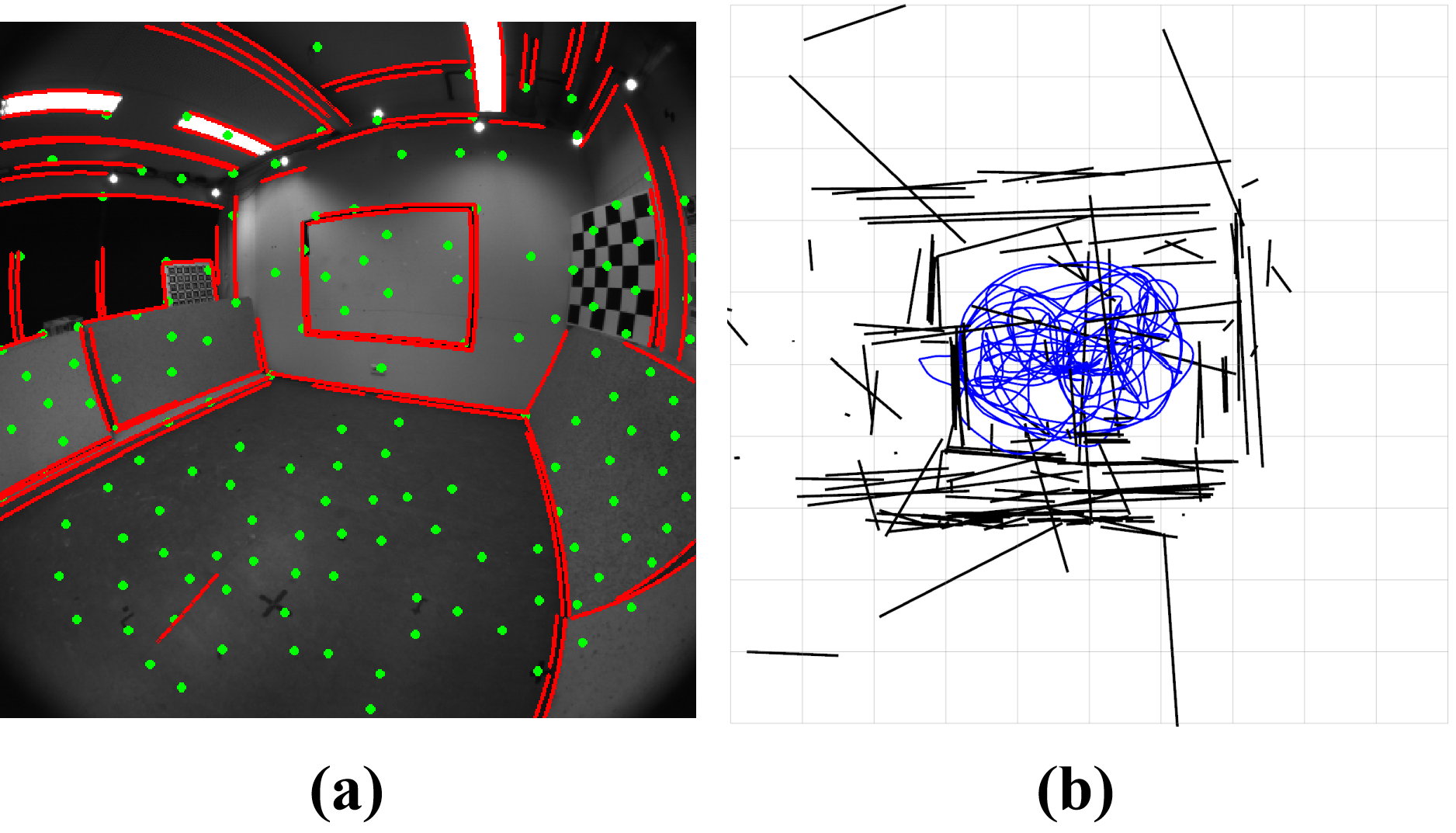}
	%\vskip-1ex
        \caption{Results of LF-PGVIO on the TUM Room1 dataset~\cite{schubert2018tum}: $\textbf{(a)}$ Point and curve segments detection. $\textbf{(b)}$ Line map results and trajectories.}
	\label{fig:TUM_vis}
        \vskip-3ex
\end{figure}

Thus, it verifies the efficiency and robustness of our line feature residual based on geodesic segments.
The line detection and line mapping results on the TUM Room1 dataset are shown in Fig.~\ref{fig:TUM_vis}.
It can be seen that our method successfully detects most of the lines in the indoor environment with satisfactory continuity.

\noindent\textbf{Results on the PALVIO dataset}: 
The PALVIO dataset~\cite{wang2022lf} is collected with two panoramic annular cameras, an IMU sensor, and a RealSense D435 sensor.
The computing platform is a quad-core Intel i7-8550U processor.
The panoramic cameras are capable of capturing monocular images at a resolution of $1280{\times}960$ and a frame rate of $30Hz$.
They have a FoV of $360^\circ{\times}(40^\circ{\sim}120^\circ)$.
Additionally, the IMU sensor provides angular velocity and acceleration measurements at $200Hz$, while the motion capture system delivers position and attitude information at a rate of $10Hz$. 
In order to validate the performance across different environments, we have included outdoor dataset sequences OD01 and OD02, which were collected using a small vehicle, for evaluation on outdoor lawns. The ground truth for these sequences was calculated using a Livox Mid-360 LiDAR sensor by Fast-LIO2~\cite{xu2022fast}, as shown in Fig.~\ref{fig:car}.
For the experiment on the PALVIO dataset, we set the maximum number of feature points in an image to $250$ and use a sliding window size of $10$.

\begin{figure}[t]
	\centering
	\includegraphics[width=1.0\linewidth]{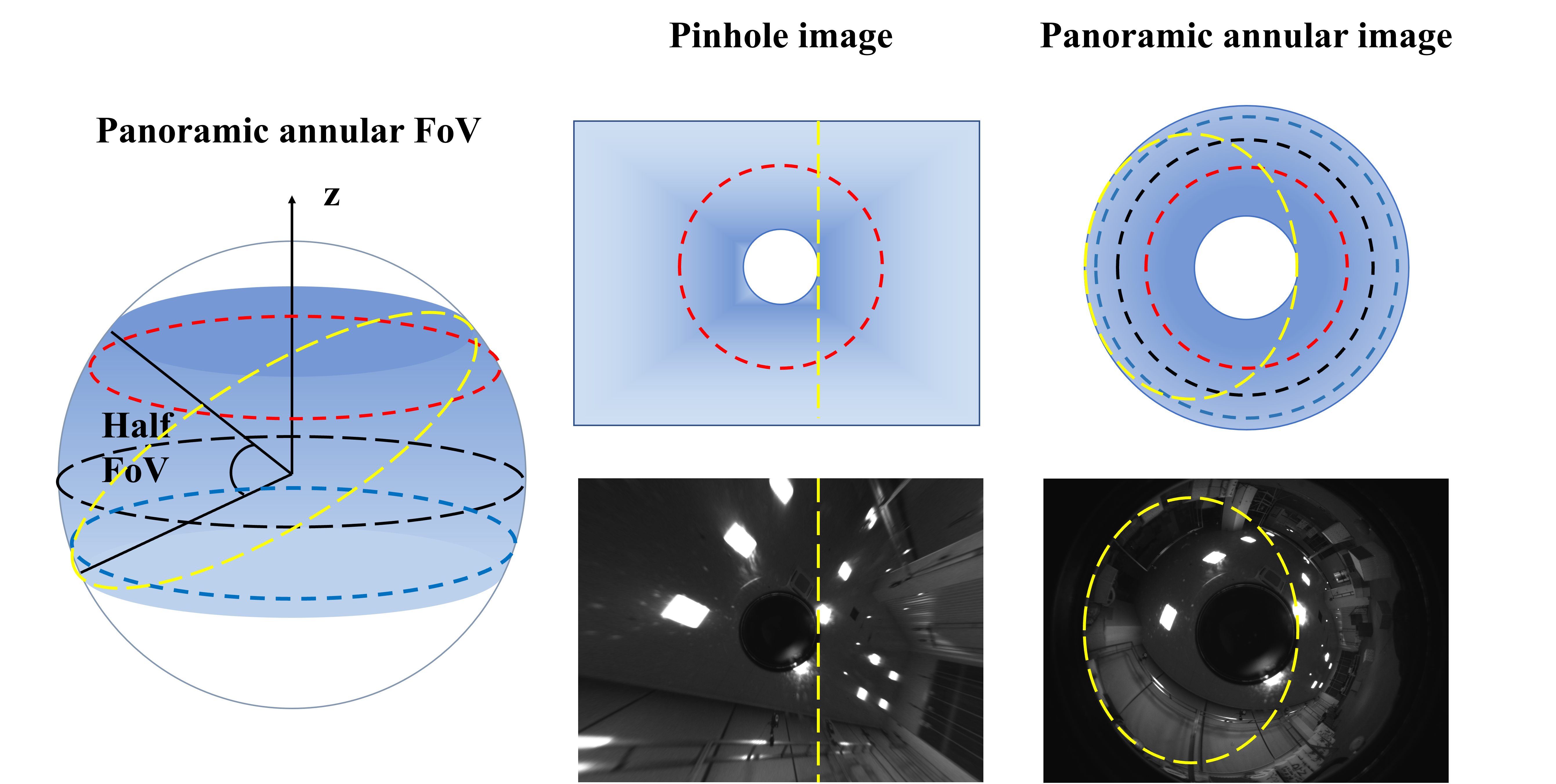}
	% \vskip-1ex
        \caption{{The yellow line represents the great circle of the unit sphere, which is projected in the pinhole image and the panoramic annular image as shown on the right.}}
	\label{fig:pinhole}
        \vskip-3ex
\end{figure}

We compare our proposed LF-PGVIO with state-of-the-art LF-VIO~\cite{wang2022lf}, SVO2.0~\cite{forster2014svo}, VINS-Mono~\cite{qin2018vins}, PL-VINS~\cite{fu2020pl}, and UV-SLAM~\cite{lim2022uv} in Table~\ref{tab:lf-pgvio}. 
For the PL-VINS and UV-SLAM algorithms, as the line features of these two methods are required to be straight lines on the image to be detected, we expand part of the positive half-plane into a pinhole image and then extract the line features for comparison in Fig.~\ref{fig:pinhole}.
To facilitate fair comparisons, we use the camera model introduced by Scaramuzza~\textit{et al.}~\cite{scaramuzza2006toolbox} for point feature extraction in all algorithms. 
For line feature extraction, LF-PGVIO employs the model from Scaramuzza~\textit{et al.}, while other methods use the pinhole camera model. 
All experiments are conducted on a laptop with an R7-5800H processor.
We employ RPEt, RPEr, and ATE as the metrics to evaluate the VIO systems.

\begin{figure*}[t!]
	\centering
	\subfigure[{\tiny Top Trajectory on ID01}]{
		\includegraphics[width=0.22\textwidth]{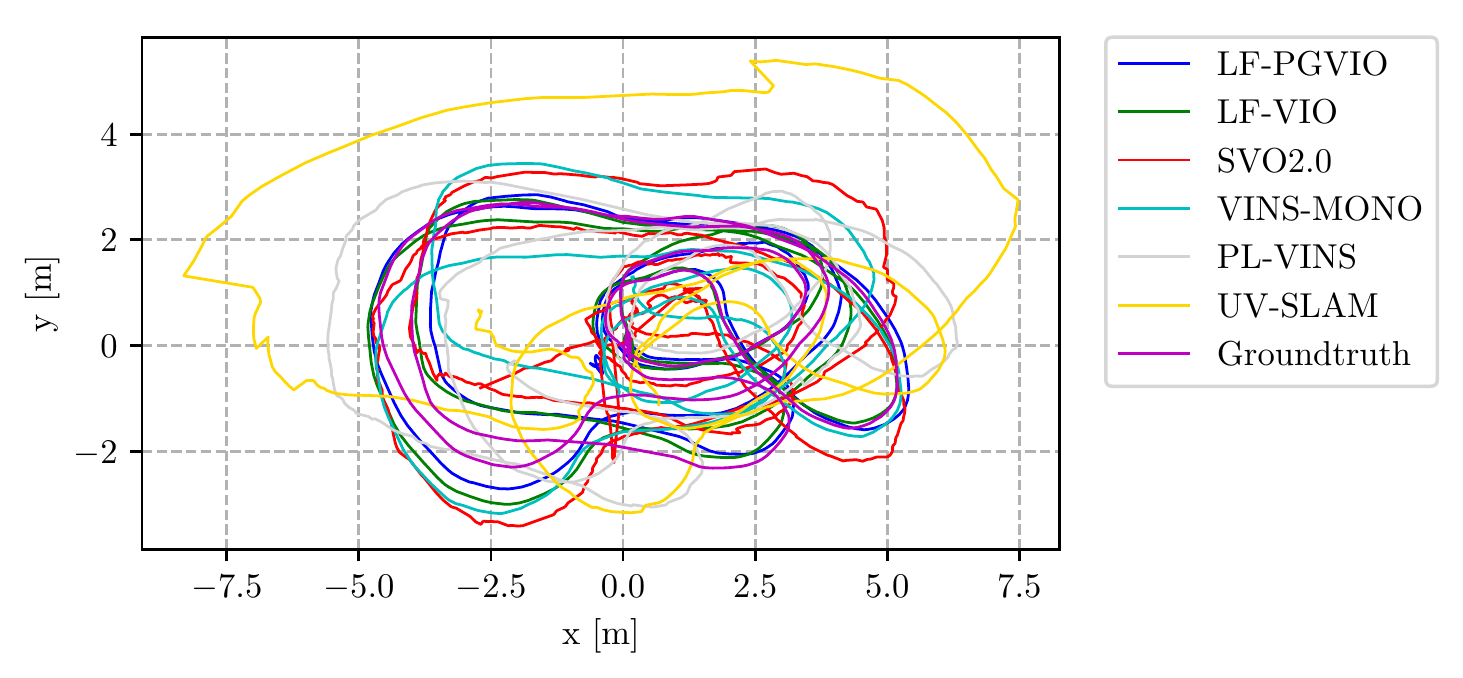}
		\label{fig:mh_01_trajectory_top}
	}
	\subfigure[{\tiny Top Trajectory on ID06}]{
		\includegraphics[width=0.22\textwidth]{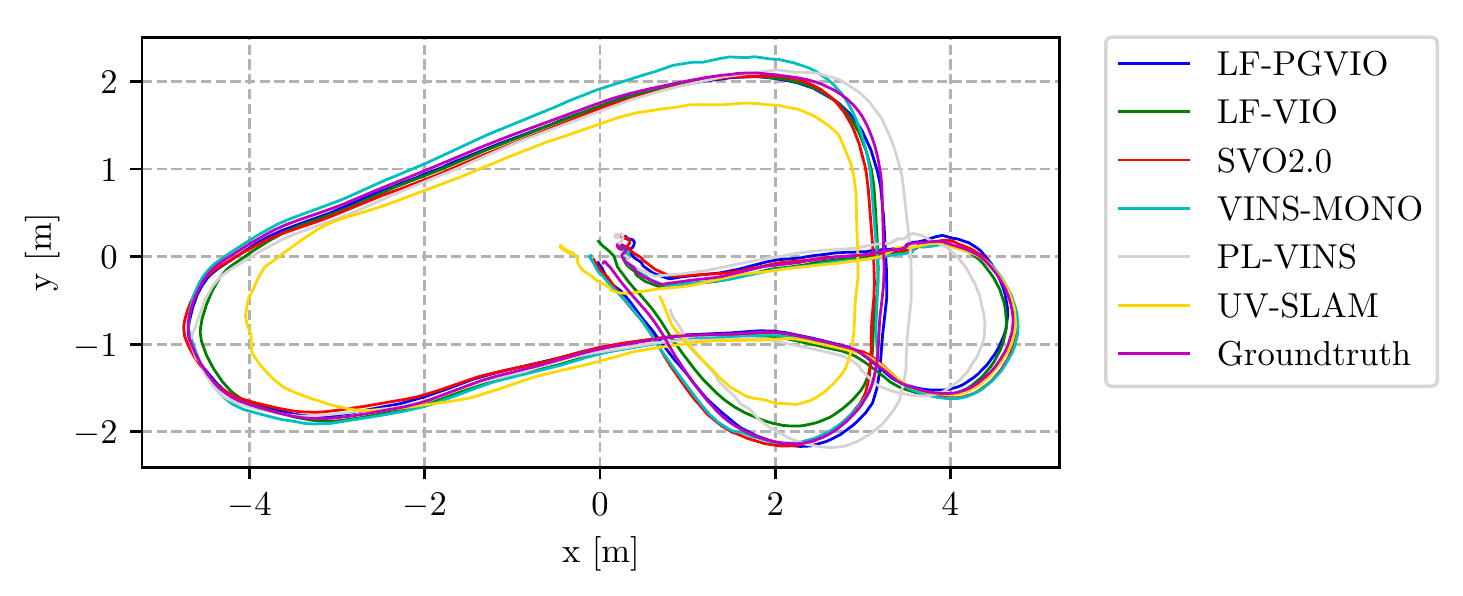}
		\label{fig:mh_06_trajectory_top}
	}
	\subfigure[{\tiny Top Trajectory on ID10}]{
		\includegraphics[width=0.22\textwidth]{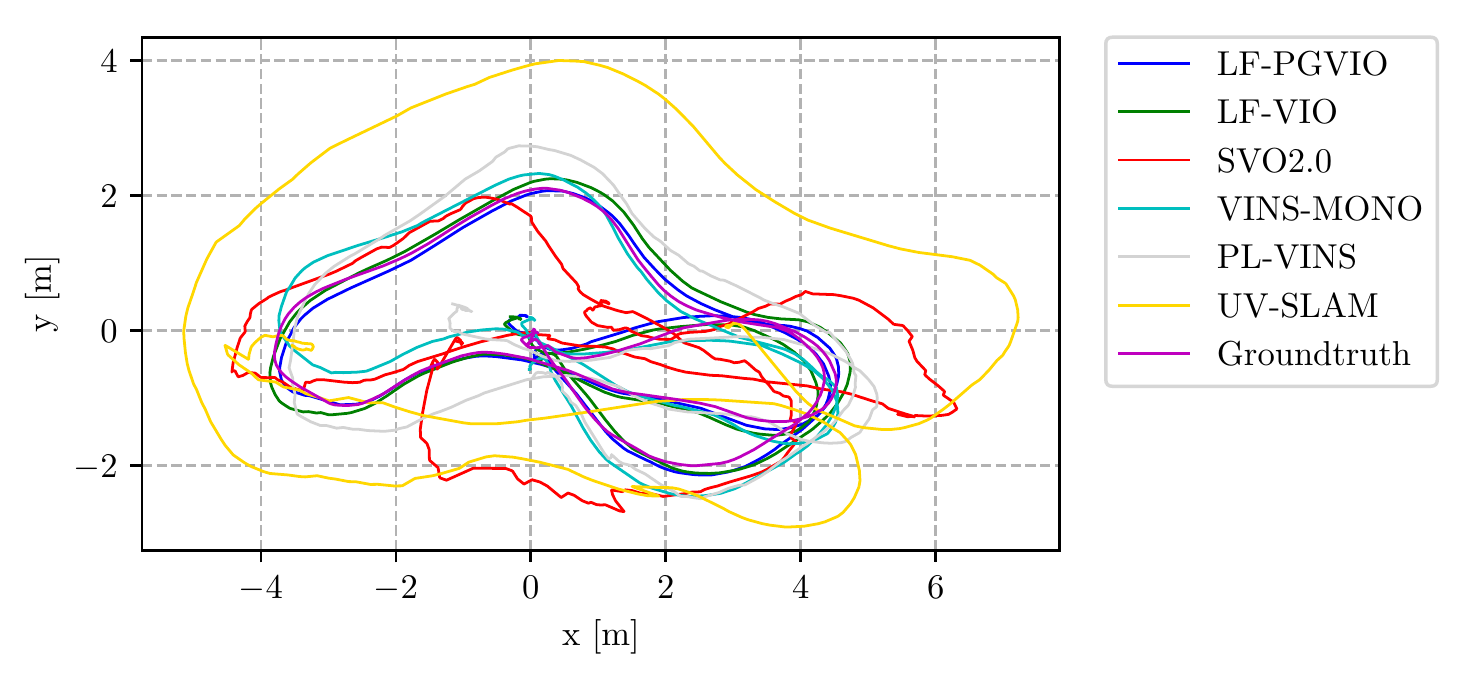}
		\label{fig:mh_11_trajectory_top}
	}
 	\subfigure[{\tiny Top Trajectory on 0D01}]{
		\includegraphics[width=0.22\textwidth]{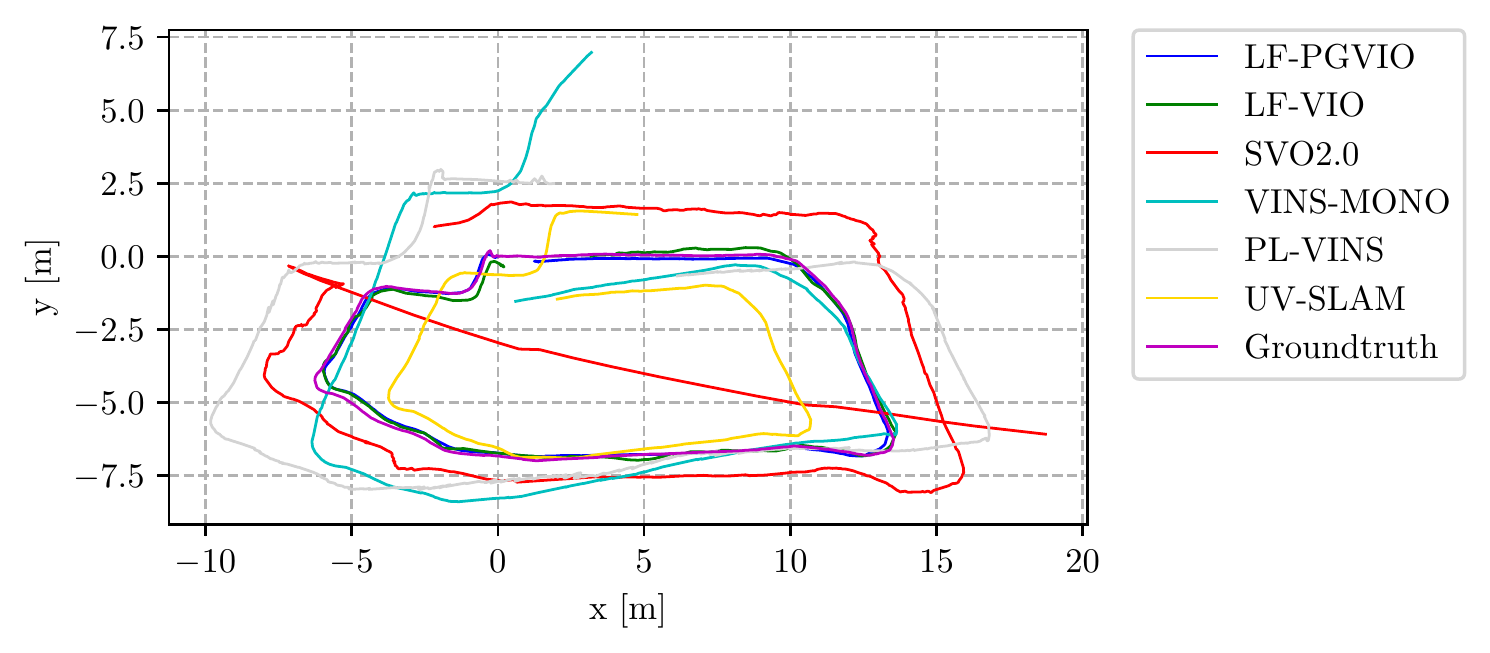}
		\label{fig:od_01_trajectory_top}
	}
	\subfigure[{\tiny Translation and Rotation Error on ID01}]{
		\includegraphics[width=0.22\textwidth]{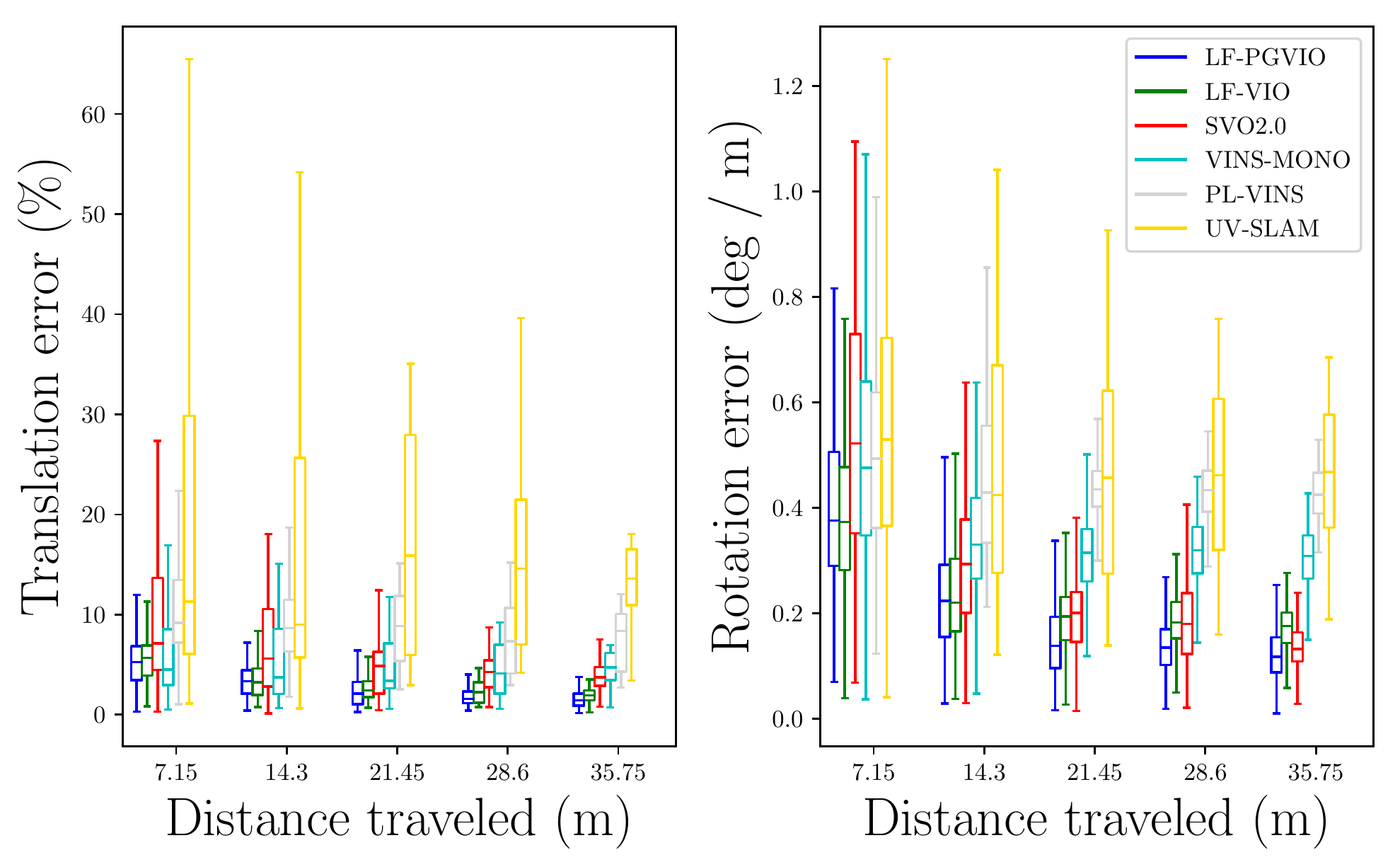}
		\label{fig:mh_01_trans_rot_error}
	}
	\subfigure[{\tiny Translation and Rotation Error on ID06}]{
		\includegraphics[width=0.22\textwidth]{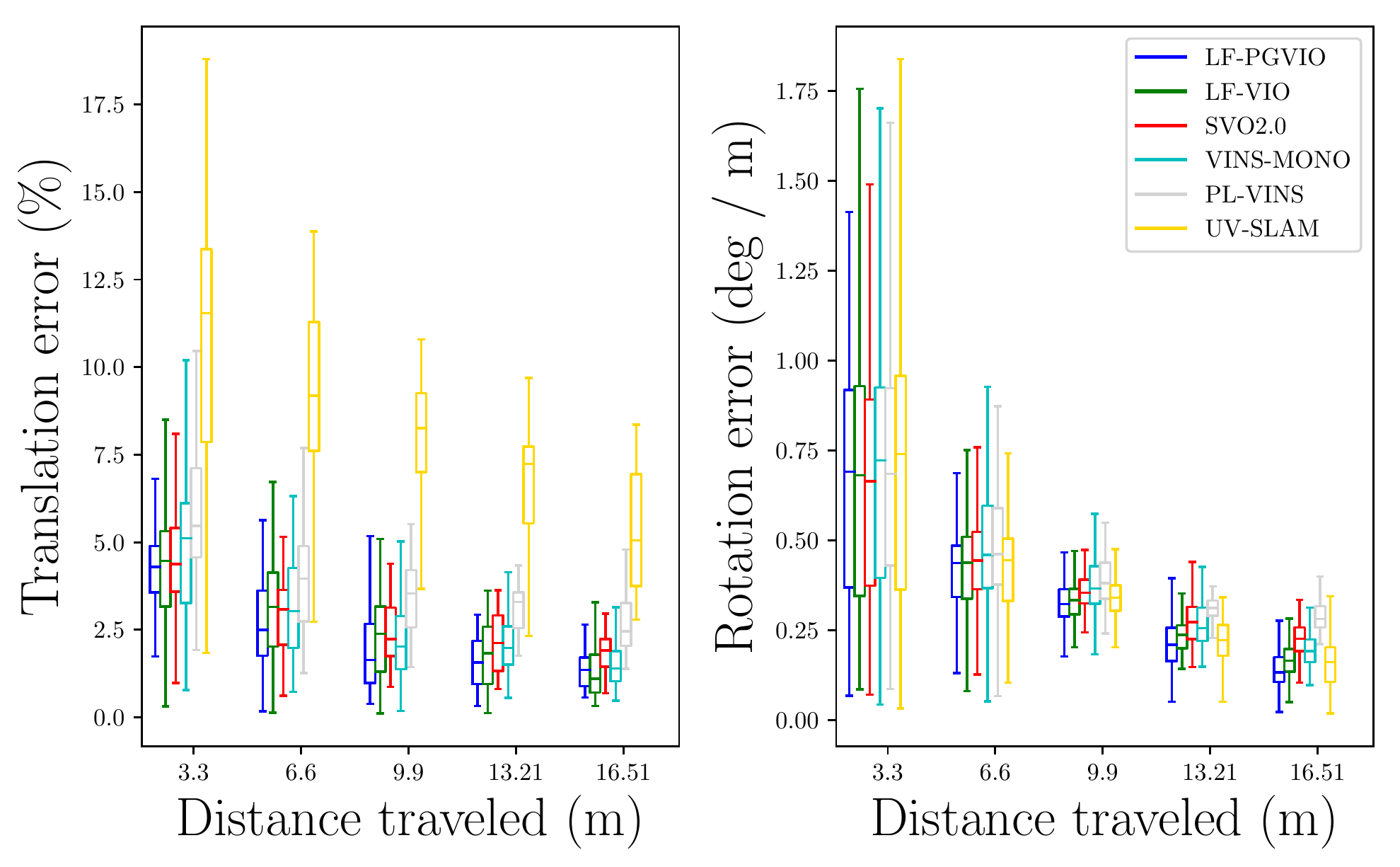}
		\label{fig:mh_06_trans_rot_error}
	}
	\subfigure[{\tiny Translation and Rotation Error on ID10}]{
		\includegraphics[width=0.22\textwidth]{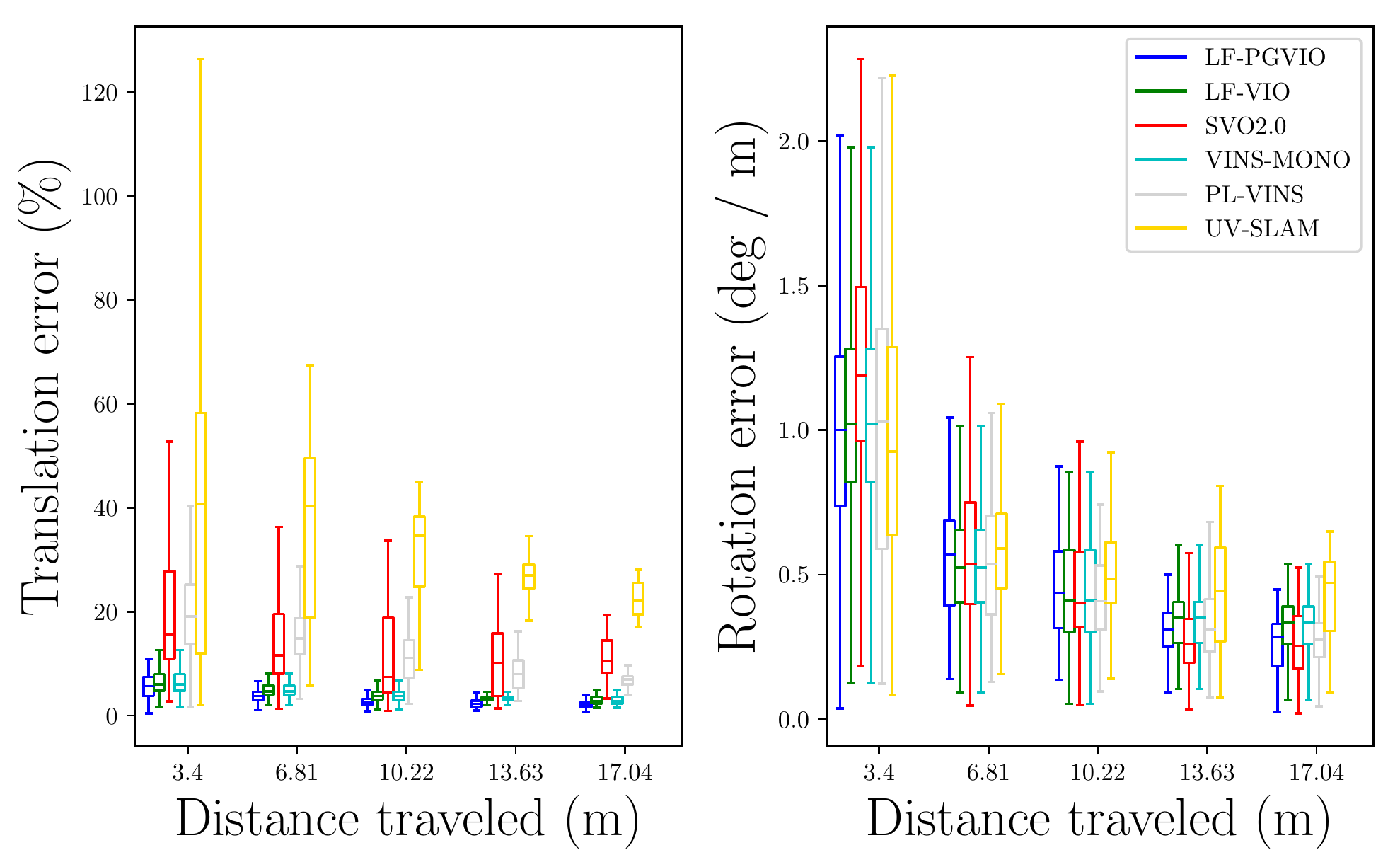}
		\label{fig:mh_11_trans_rot_error}
	}
 	\subfigure[{\tiny Translation and Rotation Error on OD01}]{
		\includegraphics[width=0.22\textwidth]{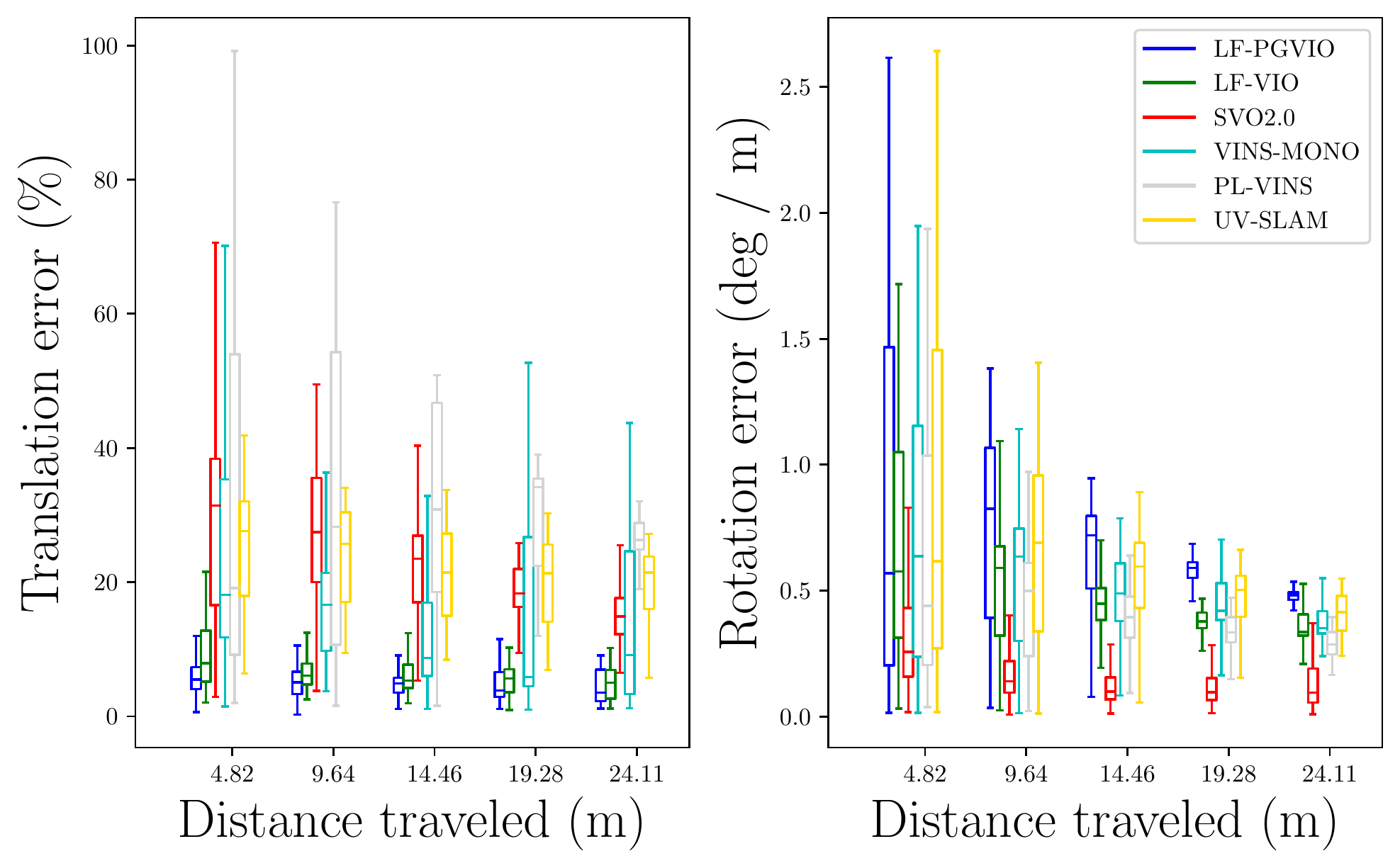}
		\label{fig:od_01_trans_rot_error}
	}
	\vskip-1ex
	\caption{Examples of top trajectories and error analyses of different VIO systems on the PALVIO dataset~\cite{wang2022lf}.}
	\label{fig:lfvio-vins-svo}
	\vskip-1ex
\end{figure*}

\begin{figure}[t!]
	\centering
	\includegraphics[width=0.85\linewidth]{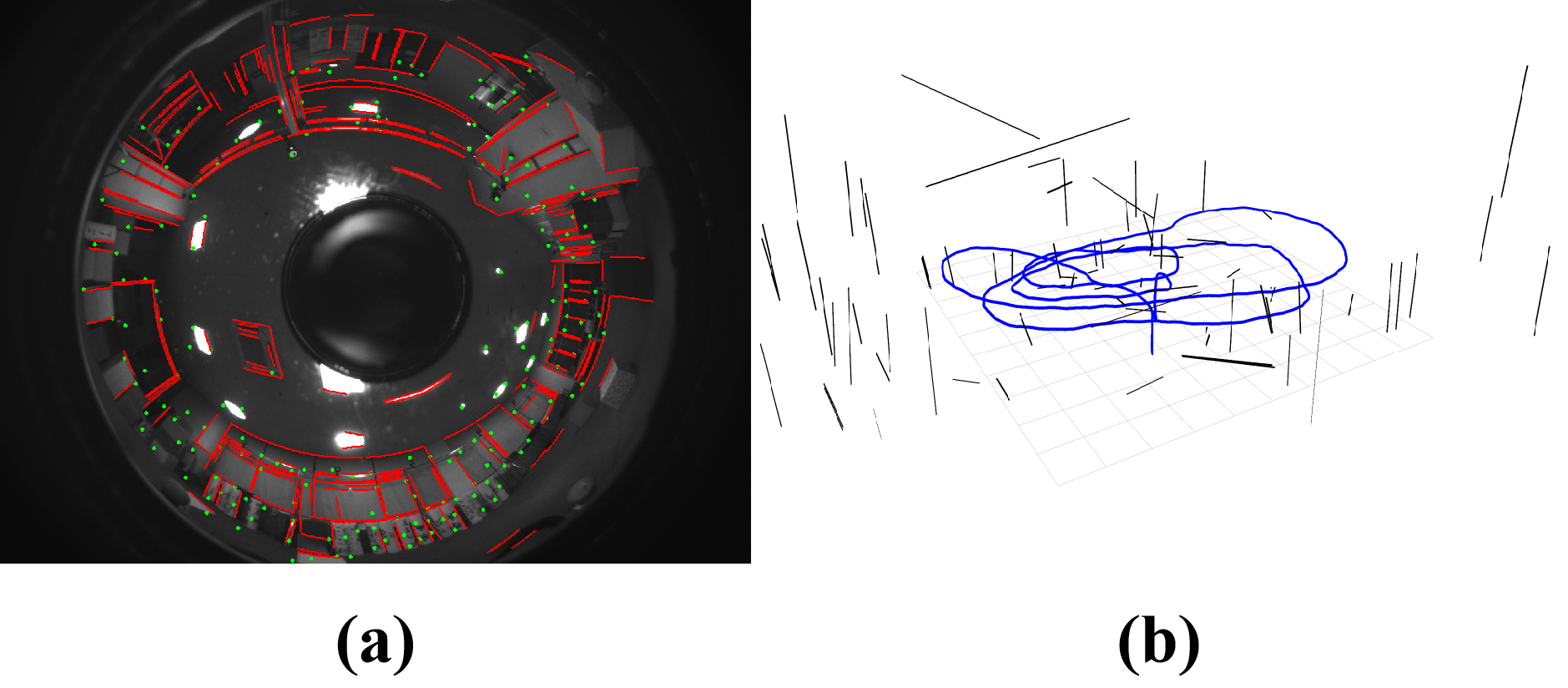}
	\vskip-1ex
        \caption{Results of LF-PGVIO on the PALVIO ID01 sequence~\cite{wang2022lf}: $\textbf{(a)}$ Point and curve segments detection in the raw image. $\textbf{(b)}$ Line map results and trajectory.} 
	\label{fig:PALVIO_vis}
 \vskip-3ex
\end{figure}

In Table~\ref{tab:lf-pgvio} and Fig.~\ref{fig:lfvio-vins-svo}, it can be observed that LF-PGVIO delivers more accurate performance in terms of RPEt, RPEr, and ATE in most sequences.
The line detection visualization and line mapping results using the ID01 sequence of PALVIO are shown in Fig.~\ref{fig:PALVIO_vis}. 
As shown in Fig.~\ref{fig:pinhole}, if the center of the pinhole image is back-projected directly above the unit sphere, the great circle with the black dotted line cannot be observed, and the great circle with the yellow dotted line may be cut off. 
However, our method can directly detect curve segments on the original image and ensure that the curve segments are not cut off if there is a curve segment crossing the negative half-plane, as shown in Fig.~\ref{fig:PALVIO_vis}.
It can be seen that our method can detect most of the lines in the indoor environment with satisfactory continuity. The top view of the OD01 sequence in outdoor experiments is shown in Fig.~\ref{fig:car}. 
For outdoor sequences, the performance of LF-PGVIO and LF-VIO has obvious advantages over other algorithms, and the accuracy of LF-PGVIO is higher than that of LF-VIO. 
However, the other four algorithms show unsatisfactory robustness when dealing with large field-of-view cameras. 
Additionally, the uneven terrain of outdoor lawn environments can cause some bumps during car operations. 
This leads to VINS-Mono and UV-SLAM failing in the outdoor OD02 sequence.
In summary, for large-FoV images, the line constraints constructed by geodesic segments help greatly enhance the accuracy and robustness of the VIO system.

\begin{figure}[t!]
	\centering
	\includegraphics[width=1.0\linewidth]{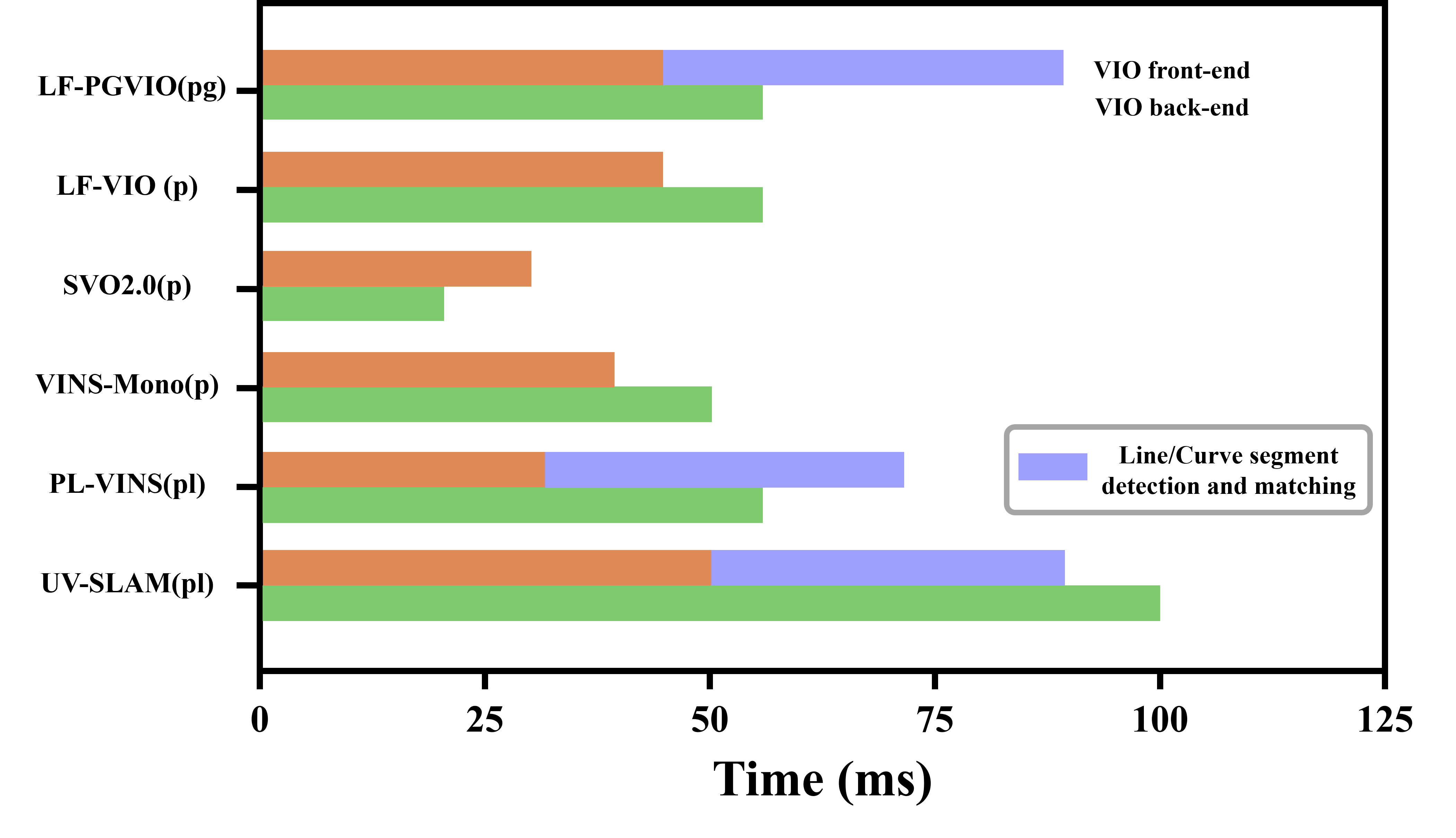}
	\caption{Elapsed time compared with different VIO methods.}
	\label{fig:speed_analysis}
    \vskip-3ex
\end{figure}

\subsection{Speed Analysis}
In this section, we present the efficiency analysis using the TUM dataset with an input size of $512{\times}512$ and the PALVIO dataset with an input size of $1280{\times}960$, with an onboard computer equipped with a quad-core Intel i7-8550U processor.
The front-end and back-end modules operate independently on two threads. 
For the TUM dataset, it takes ${\sim}50ms$ and ${\sim}50ms$, respectively. In the front end, the feature detection and optical flow computation cost $3ms$ and $1ms$. The curve segments detection and matching take $25ms$ and $0.2ms$. In the back end, the solver takes $35ms$. 
For the PALVIO dataset, it takes ${\sim}80ms$ and ${\sim}55ms$, respectively. In the front end, the feature detection and optical flow computation cost $15ms$ and $3ms$. The curve segments detection and matching take $50ms$ and $0.4ms$. In the back end, the solver takes $35ms$. 
The elapsed time of LF-PGVIO compared with different VIO methods is shown in Fig.~\ref{fig:speed_analysis}. Among them, PL-VINS and UV-SLAM use line segment detection and matching, and our method LF-PGVIO uses curve segment detection and matching. While the FoV of the panoramic image in LF-PGVIO is larger, our method maintains efficiency and the segment detection and matching time is close to that of PL-VINS and UV-SLAM.
 
Overall, LF-PGVIO achieves a frame rate of at least $10Hz$ on the onboard computer and is reasonable for real-time mobile applications on intelligent vehicles and robots. If the odometry is used for control, the IMU integral data is usually employed to generate the odometry based on the $10Hz$ optimized output. The frequency of this odometry depends on the IMU frequency, which in our dataset is $200Hz$. 
In summary, our method is ideally suitable for mobile applications in intelligent vehicles and robots.

\section{Conclusion}
In this paper, we present a novel point-line visual-inertial odometry (VIO) framework with omnidirectional cameras. By incorporating points and geodesic segments and extending line feature residuals, our method unleashes the potential of point-line odometry with large-FoV omnidirectional cameras, even for cameras with negative-plane FoV.

The LF-PGVIO framework demonstrates high versatility and can be applied to various camera models. It exhibits excellent performance in large-FoV camera models, surpassing state-of-the-art VIO approaches in terms of robustness and accuracy. The potential impact of our findings on large-FoV VIO is significant, with wide-ranging potential applications in fields such as robotics and autonomous navigation.

However, there is still scope for enhancing LF-PGVIO. Although curve segment detection methods exhibit high accuracy and fast detection time, their limitations arise when employed for loop closure tasks due to the effectiveness of curve segment matching methods being restricted to images with smaller parallax.

In the future, we are interested in overcoming the problem of decreased line matching rate caused by perspective changes by utilizing learning methods or geometric characteristics, as well as adding loop closure using line features and plane constraints to our system to further improve localization accuracy.

%%%%%%%%% REFERENCES
{\small
\bibliographystyle{IEEEtran}
\bibliography{bib}
}

\end{document}